\documentclass[lettersize,journal]{IEEEtran}
\usepackage{amsmath,amsfonts}
\usepackage{algorithmic}
\usepackage{algorithm}
\usepackage{array}
\usepackage[caption=false,font=normalsize,labelfont=sf,textfont=sf]{subfig}
\usepackage{textcomp}
\usepackage{stfloats}
\usepackage{url}
\usepackage{verbatim}
\usepackage{graphicx}
\usepackage{framed,multirow}
\def\BibTeX{{\rm B\kern-.05em{\sc i\kern-.025em b}\kern-.08em
    T\kern-.1667em\lower.7ex\hbox{E}\kern-.125emX}}
\usepackage{balance}
\usepackage{biblatex}
\usepackage[table]{xcolor}
\usepackage{amsmath}
\usepackage{graphicx}
\usepackage{csquotes}
\usepackage{booktabs} 
\usepackage{amssymb}
\usepackage{latexsym}
\usepackage{hyperref}
\usepackage{framed,multirow}
\usepackage{booktabs}
\begin{document}
\title{USIGAN: \underline{U}nbalanced \underline{S}elf-\underline{I}nformation Feature Transport for Weakly Paired  Image IHC Virtual Staining}
\author{Yue Peng$^ \dag$,Bing Xiong$^ \dag$,Fuqiang Chen,Deboch Eybo Abera, RanRan Zhang, Wanming Hu,Jing Cai,Wenjian Qin$^ *$
\thanks{$\dag$ Yue Peng and Bing Xiong contribute equally to this work. }
\thanks{$*$ Wenjian Qin are corresponding author to this work. }
\thanks{B. Xiong, Y. Peng, F.Chen, Deboch. Abera, R. Zhang are with the ShenZhen Institues of Advanced Technology, university chinese academy of sciences , ShenZhen,China, 518055. Email: wj.qin@siat.ac.cn.}
\thanks{W. Hu are with Department of Pathology, Sun Yat-sen University Cancer Center, State Key Laboratory of Oncology in South China,Guangdong Provincial Clinical Research Center for Cancer, Guangzhou 510060, China}
\thanks{J.Cai are with Department of Health Technology and Informatics, The Hong Kong Polytechnic University, Hong Kong 999077, China}
}

\markboth{IEEE TRANSACTIONS ON IMAGE PROCESSING}%
{How to Use the IEEEtran \LaTeX \ Templates}

\maketitle

\begin{abstract}
Immunohistochemical (IHC) virtual staining is a task that generates virtual IHC images from H\&E images while maintaining pathological semantic consistency with adjacent slices. This task aims to achieve cross-domain mapping between morphological structures and staining patterns through generative models, providing an efficient and cost-effective solution for pathological analysis. However, under weakly paired conditions, spatial heterogeneity between adjacent slices presents significant challenges. This can lead to inaccurate one-to-many mappings and generate results that are inconsistent with the pathological semantics of adjacent slices.
To address this issue, we propose a novel unbalanced self-information feature transport for IHC virtual staining, named USIGAN, which extracts global morphological semantics without relying on positional correspondence. By removing weakly paired terms in the joint marginal distribution, we effectively mitigate the impact of weak pairing on joint distributions, thereby significantly improving the content consistency and pathological semantic consistency of the generated results. 
Moreover, we design the Unbalanced Optimal Transport Consistency Mining (UOT-CTM) mechanism and the Pathology Self-Correspondence Mining (PC-SCM) mechanism to construct correlation matrices between H\&E and generated IHC in image-level and  real IHC and generated IHC image sets in intra-group level.
Experiments conducted on two publicly available datasets demonstrate that our method achieves superior performance across multiple clinically significant metrics, such as IoD and Pearson-R correlation, demonstrating better clinical relevance. The code is available at: \url{https://github.com/MIXAILAB/USIGAN}
\end{abstract}


\section{Introduction}
\IEEEPARstart{I}{mmunohistochemistry}(IHC)  virtual staining is a task that aims to generate virtual IHC images consistent with the pathological semantics of adjacent reference IHC-stained slices while preserving the content consistency of H\&E images. However, spatial heterogeneity between adjacent slices makes it challenging to directly establish a mapping between H\&E and IHC images. In previous studies, most researchers adopted contrastive learning approaches, which maximize mutual information in the corresponding regions to establish cross-domain correlations between the morphological information in H\&E images and the staining styles in IHC images. However, this approach still fails to address the issue of spatial heterogeneity, which can lead to misalignment of morphological information from H\&E images to incorrect IHC staining patterns, affecting the extraction and accurate mapping of key information in generative models~\cite{ref_2,ref_3,ref_10}. Therefore, we focus on a previously overlooked concept in information theory—self-information, which quantifies the information content of a single event. Specifically, self-information reflects the rarity of a pixel. For example, when rare structures or patterns appear in pathological images, the corresponding regions exhibit higher self-information, indicating that they contain more valuable diagnostic information and biological significance. In summary, self-information quantifies the model's prioritization of information needs for critical pathological features.

\begin{figure}[!t]
\centerline{\includegraphics[width=\columnwidth]{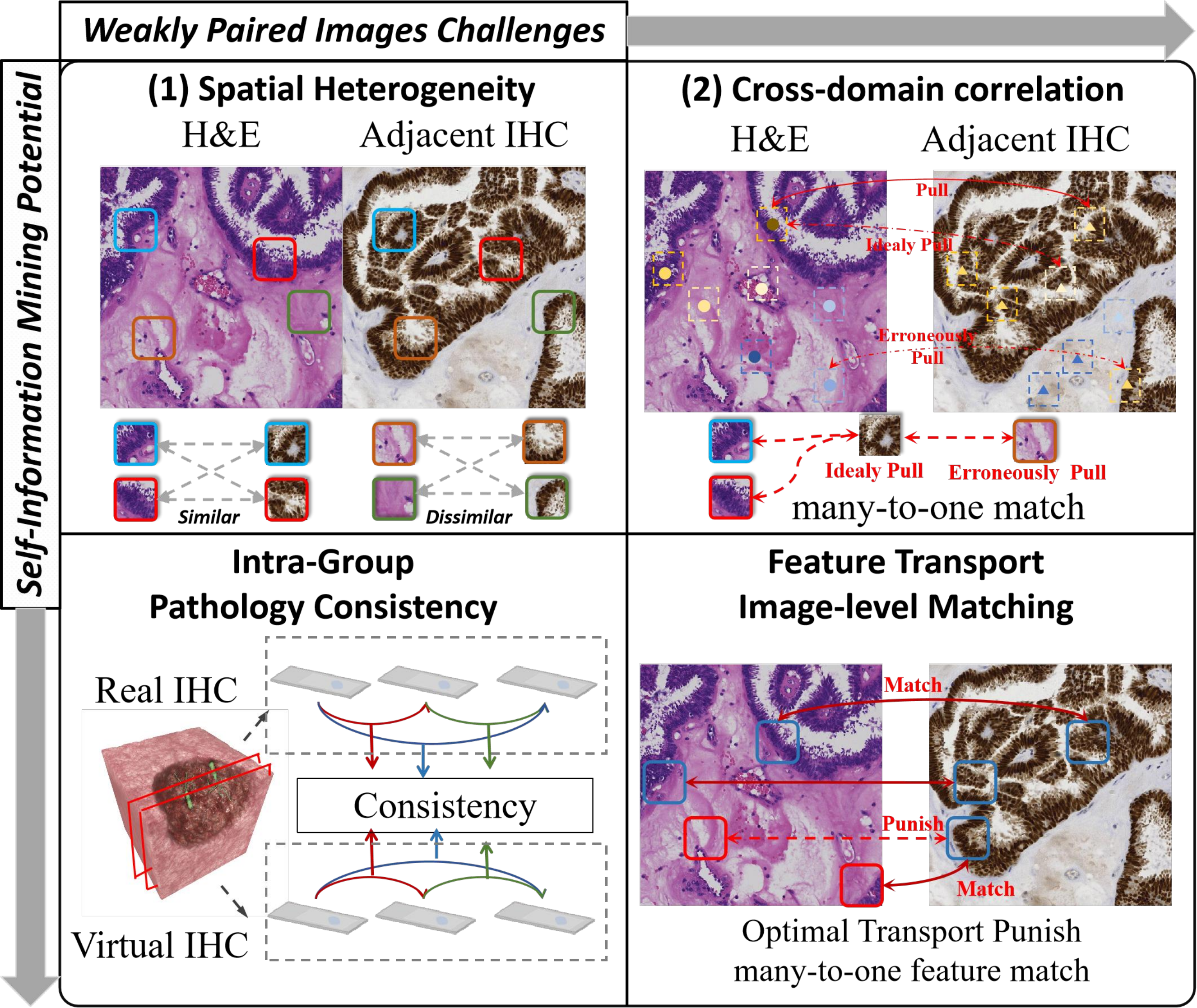}}
\caption{Spatial heterogeneity between adjacent slices poses a major challenge in virtual staining. This heterogeneity manifests as misalignment of tissue structures and pathological semantics across slices, making it difficult to directly establish cross-domain mappings between morphology and staining styles.}
\label{fig1}
\end{figure}

Our goal is to transform self-information from a previously passive learning paradigm into an active focus, further uncovering the process by which generative models establish cross-domain correlations in virtual staining. Self-information represents critical regions in the mapping process, and constructing a correlation matrix allows different feature regions to be assigned higher weights, which is the essence of self-information mining.
As illustrated in Figure \ref{fig1},  IHC virtual staining faces two main challenges: (1) spatial heterogeneity and (2) erroneous pathological semantic correlations caused by incorrect one-to-many mappings between morphological features and staining patterns. Self-information mining addresses these challenges by prioritizing key information.
We leverage two types of priors in virtual staining to incorporate corresponding self-information and establish correlations:
1) Explicit Prior: Pathological semantic consistency between adjacent slices. By evaluating staining intensity through clinical optical density metrics, we construct a correlation matrix between generated IHC and real IHC image groups to mine pathological semantic self-information.
2) Implicit Prior: Morphological structures and cellular atypia in H\&E images correspond to different disease subtypes and their respective IHC staining patterns~\cite{ref_49}. Through optimal transport theory, we optimize the expectation of self-information, explicitly introducing self-information to strengthen critical regions and penalizing the effects of erroneous one-to-many mappings.

In previous studies on IHC virtual staining, two paradigms: fully supervised and unsupervised—have been primarily used to passively construct cross-domain correlations. Pix2Pix and its variants attempt to establish cross-domain correlations at the pixel level by leveraging weakly paired IHC data~\cite{ref_14,ref_15,ref_16,ref_17}. However, these methods struggle to effectively learn mapping relationships between tissue regions. Cycle consistency-based methods implicitly construct mappings between morphology and staining styles by allowing two models to map between two domains~\cite{ref_28}. However, the assumption of reciprocal mappings between staining domains is often violated for weakly paired data, resulting in low pathological semantic consistency in the generated results. To address these issues, some researchers have adopted contrastive learning approaches that maximize mutual information between patches to explicitly construct mapping relationships~\cite{ref_14,ref_15,ref_16,ref_17}. While this non-pixel-level approach alleviates the impact of weakly paired data to some extent, its performance remains constrained by the degree of pairing. These methods overlook the critical role of self-information in pathological images. Passively summarizing mapping relationships from image collections is inherently challenging, whereas self-information, by emphasizing features with high diagnostic value, can guide the model to correctly focus on key regions, thereby improving the performance and consistency of virtual staining results.

In this paper, we propose a novel method for mining self-information at both the image and group levels by eliminating weakly paired terms in the joint marginal distribution, effectively mitigating their impact and significantly enhancing the content and pathological semantic consistency of generated results. Specifically, for image-level self-information mining, we introduce Unbalanced Optimal Transport Consistency Mining (UOT-CTM), a method that preserves geometric structures of high self-information regions to reduce transport costs during optimal transport minimization. Although previous attempts using optimal transport explicitly constrained transport plans from H\&E to weakly paired IHC and from generated IHC to weakly paired IHC in feature space—thus aligning morphological and staining style features—these approaches suffer from three primary limitations: 1) Neglecting incompatibility in weakly paired data, leading to forced, low-similarity feature matching. 2) Ignoring inherent semantic differences between morphological and staining-style imitation tasks, resulting in undesirable trade-offs. 3) Overlooking semantic inconsistencies and distributional differences by directly constraining joint marginal distributions involving weakly paired terms.
To address these issues, UOT-CTM employs a cyclic consistent transport strategy to explicitly constrain transport plans along paths from H\&E to generated IHC, effectively capturing image-level self-information. Additionally, we propose a marginally relaxed unbalanced optimal transport strategy to alleviate the one-to-many mapping issue and construct more accurate correlation matrices, transforming the constraint objective from directly matching \(P(x, y)\) and \(P(y, z)\) distributions to matching \(P(x, z)\) and \(P(x, y) \cdot P(y, z)\), thus eliminating weakly paired terms. For group-level self-information mining, we design the Pathology-Consistent Self-Correspondence Mining (PC-SCM) mechanism, which leverages optical density—rare but indicative of high self-information—as a ``self-information anchor,'' establishing correlation matrices between generated and real IHC based on optical density. This ensures prioritization of high-optical-density regions, maintains consistent correlation matrices within batches, avoids negative impacts from directly using weakly paired IHC as learning templates, and effectively captures pathological semantic information. Our contributions can be concluded as follows:

\begin{itemize}
    \item We introduce the concept of self-information of the image in histopathology. By maximizing this self-information, we achieve a marked boost in cross-domain correlation accuracy, delivering virtual IHC results that are superior in both content and pathological semantic consistency. 
    \item To mine  image-level self-information  we employ unbalanced optimal transport, which mitigates the adverse effects of flawed one-to-many mappings through feature-matching transport.
    \item  {We utilize optical density to directly anchor pathological semantic information, guiding intra-batch self-information mining. This approach focuses on abnormal regions represented by high optical density, thereby indirectly leveraging self-information and mitigating the impact of spatial heterogeneity.}
\end{itemize}

\section{Related Work}
\subsection{Image-to-Image Translation}

The task of image translation aims to map images from a source domain to a target domain while preserving their original content~\cite{ref_26,ref_27,ref_40,ref_41}. Pix2Pix used a patch-based discriminator on paired data to alleviate the tendency of previous generative models toward style averaging~\cite{ref_14}, establishing itself as a cornerstone method in supervised image translation. Subsequent supervised methods introduced various extensions, such as cross-layer connections for high-resolution generation and perceptual constraints~\cite{ref_17}. However, due to its strict requirement for pixel-level paired data, Pix2Pix is difficult to generalize to broader applications. CycleGAN~\cite{ref_28} introduced a framework based on the bidirectional mapping hypothesis and cycle consistency loss, which became the foundation for subsequent methods . %such as UGATIT~\cite{ref_29}, NICE-GAN~\cite{ref_30}, and RegGAN~\cite{ref_31}. 
To address the overly strict bidirectional mapping assumption of cycle consistency, CUT~\cite{ref_21}, based on contrastive learning, constructs cross-domain correlations using positive and negative patch samples. Contrastive learning-based image translation methods grounded in optimal transport theory have also garnered significant attention. Techniques such as weighted positive and negative samples~\cite{ref_26}, hard negative sample generation~\cite{ref_32}, and unbalanced optimal transport have been extensively studied and achieved notable progress~\cite{ref_33,ref_34}.

Although existing image translation methods show great potential in pathological image stain transfer, the lower coupling between content and style in pathological images, along with the higher demand for content consistency~\cite{ref_10}, makes it challenging for these methods to accurately generate virtual images that maintain both pathological features and content consistency.

\begin{figure*}[!htb]
\centerline{\includegraphics[width=0.9\textwidth]{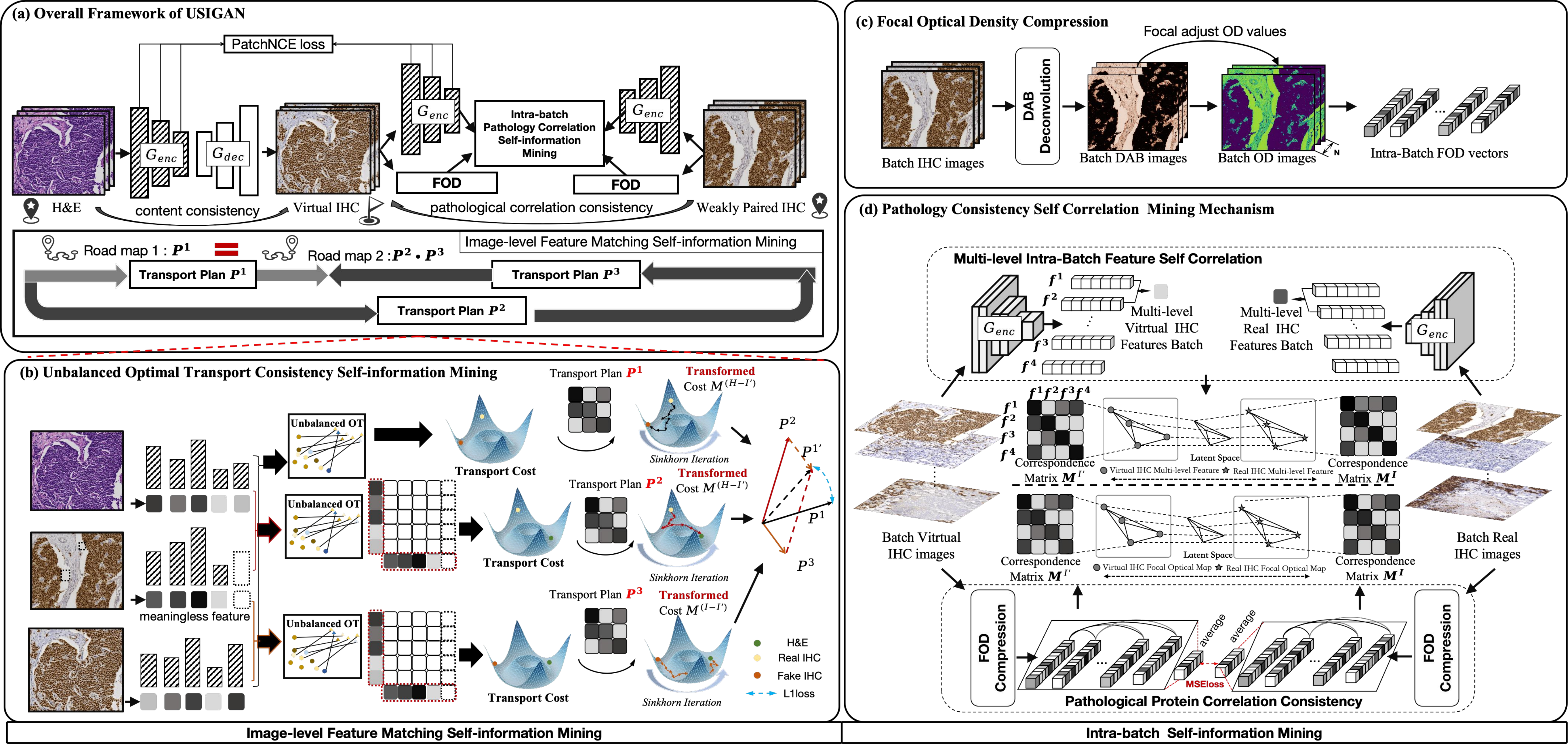}}
\caption{We leverage weakly paired IHC as an intermediate bridge to ensure global consistency between H\&E and IHC while mining image-level self-information through a transport consistency framework. Simultaneously, we utilize focal optical density and DAB deconvolution to extract optical density feature vectors to guide intra-batch self-information mining. Multi-scale features within the batch are used as auxiliary features to fully explore intra-batch self-information.}
\label{fig2}
\end{figure*}

\subsection{Immunohistochemical Virtual Staining}
Immunohistochemical virtual staining aims to generate staining results for specific protein markers from source-stained images while preserving image content. This technique is gaining increasing attention in digital pathology. Zhang et al. proposed an IHC multi-staining mapping method that measures the accuracy of specific protein markers based on mean intensity optical density (MIOD), introducing a novel evaluation metric for assessing the molecular response accuracy of virtual staining results~\cite{ref_35}. Liu et al. proposed a method that annotates positive regions in breast tissue H\&E images to achieve high pathological correlation with adjacent slice labels, further demonstrating that morphological information can effectively distinguish positive regions~\cite{ref_4}. Chen et al., on the other hand, sought to reduce the demand for labeled positive signal data by using DAB deconvolution for destaining, segmenting positive regions based on optical density thresholds, and employing an additional feature extractor to refine pseudo-labels~\cite{ref_24}, thereby achieving high preservation of pathological semantics. However, this method heavily relies on the quality of the feature extractor, which limits its practicality. It is evident that the morphological information in H\&E images plays a crucial role in maintaining pathological relevance.

{At present, the ability of optimal transport to penalize many-to-one mappings has been applied by researchers in weakly paired or unpaired image translation tasks~\cite{ref_26,ref_ufte}. Guan et al. proposed leveraging optimal transport to compute the transport cost between H\&E and virtual IHC, as well as between weakly paired IHC and virtual IHC, imposing an L1 constraint in the feature space~\cite{ref_25,guan2025ot}. This approach encourages virtual IHC to achieve a balance between content consistency and pathological semantics across H\&E and weakly paired IHC. However, due to its disregard of the classical OT assumptions and failure to fully eliminate the influence of weakly paired terms, the method produces suboptimal results.}

\section{Method}
% -----缩字-----返稿
% In this chapter, we designed the image-level self-information mining strategy UOT-CTM and the batch-level self-information mining strategy PC-SCM. We introduced the optimal transport framework and demonstrated that weakly paired H\&E-IHC data does not satisfy the assumptions of classical optimal transport. Additionally, optical density, a clinical metric for evaluating staining intensity and abnormal molecular expression, was utilized as a self-information anchor to guide batch-level self-information mining. Our method eliminates weakly paired terms in the joint marginal distribution, significantly mitigating the impact of spatial heterogeneity caused by weakly paired data.

\begin{figure}[!htb]
    \centering
        \includegraphics[width=\linewidth]{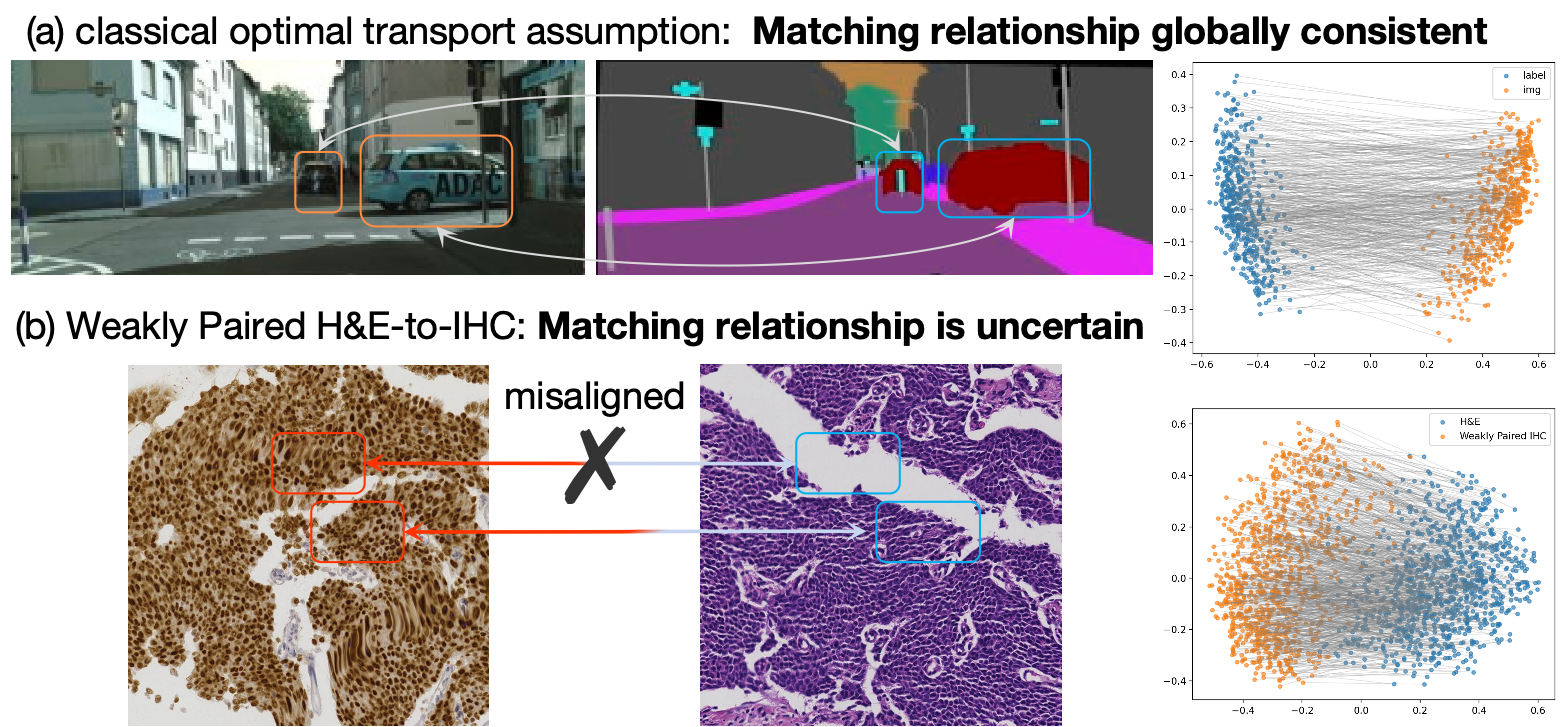}
    \caption{{Strongly paired data (Cityspace dataset) exhibit a clear global matching relationship, while weakly paired data (MIST ER) demonstrate varying matching relationships influenced by spatial heterogeneity. In features reduced by PCA, the classical optimal transport approximates a one-to-one permutation structure, which is visualized through unified open source tools POT~\cite{flamary2021pot}. The orange points in the distribution of Weakly Paired IHC may contain multiple subgroups, but the matching patterns of these subgroups do not exhibit global consistency.}}
    \label{fig:tsne}
\end{figure}

\subsection{Image-level Self-information Mining}

{Self-information serves as a measure of the information content for specific events. In pathological images, regions exhibiting the most "abnormal" characteristics are associated with high self-information. These abnormal morphological regions often correspond to areas with high expression of molecular markers. On weakly paired data, self-information provides the model with an intuitive focus on key regions. Unlike previous studies that establish cross-domain consistency by maximizing the mutual information of patches at the same spatial location, self-information reflects intra-image relationships, where abnormal regions consistently exhibit high self-information. More importantly, when cross-domain correlations are established by maximizing mutual information at the same spatial location, patch-level contrastive learning can partially mitigate the impact of weakly paired data. However, under weak pairing conditions, morphological-style patches with different pathological representations may be incorrectly minimized. This many-to-one mapping relationship poses significant challenges for establishing accurate cross-domain correlations.}

{To eliminate the weakly paired terms in the optimization objective, we aim to utilize self-information, which reflects intra-image relationships, as a bridge to capture feature relationships in weakly paired images, as shown in Figure~\ref{fig2}. Optimal transport, as a method to measure the minimal cost of moving between two distributions under a given cost function, effectively penalizes many-to-one mappings. Self-information is incorporated into the dual objective of optimal transport through entropy regularization by influencing the marginal constraints and cost structure. Specifically, entropy regularization introduces an additional term into the optimal transport problem, which encourages smoother and more distributed transport plans. Self-information, as a measure of the information content within the image, is embedded into this regularization term. The Sinkhorn-Knopp algorithm leverages this entropy-regularized formulation to compute the optimal transport plan, where self-information directly impacts the gradients and the dual variables, thereby shaping the alignment process.}

{As shown in Fig.~\ref{fig:tsne}, classical optimal transport relies on the assumption of globally consistent mapping relationships, which is difficult to satisfy in weakly paired data. To construct an accurate transport cost and penalize suboptimal mappings, we propose unbalanced optimal transport to relax the strict marginal constraints. Instead of a "hard" mass conservation constraint, it introduces a "soft" penalty with a divergence measure. The detailed steps are outlined in the pseudo-algorithm~\ref{alg:algorithm}. }

\begin{algorithm}[tb]
\caption{Relaxed Sinkhorn Algorithm for Unbalanced Optimal Transport (UOT)}
\label{alg:algorithm}
\textbf{Input}: Source features $\mathbf{q} \in \mathbb{R}^{n \times I \times D}$, target features $\mathbf{k} \in \mathbb{R}^{n \times O \times D}$ \\
\textbf{Parameters}: Entropic regularization $\varepsilon>0$, relaxation $\tau>0$, iterations $\text{max\_iter}\in\mathbb{N}$ \\
\textbf{Output}: Transport plan $\mathbf{K} \in \mathbb{R}^{(n) \times I \times O}$ (batched)

\begin{algorithmic}[1]
% Cost/similarity and preprocessing
\STATE \textbf{Stage 1: Sinkhorn stage (UOT)}
\STATE Compute batched similarity (einsum):
\[
\mathbf{C}[b,i,o] \gets \sum_{d=1}^{D} \mathbf{q}[b,i,d]\;\mathbf{k}[b,o,d]
\]
\STATE $\ \mathbf{K} \gets \mathbf{C}$
\STATE Mask self-matches: $\ \mathbf{K}[b,i,i] \gets -10 \quad \forall\, b$
\STATE Entropic scaling: $\ \mathbf{K} \gets \exp\!\big(\mathbf{K}/\varepsilon\big)$
\STATE Set $(n,I,O) \gets \text{shape}(\mathbf{K})$; initialize
\[
\mathbf{u} \gets \mathbf{1}_{n\times I}, \qquad \mathbf{v} \gets \mathbf{1}_{n\times O}
\]

% Sinkhorn iterations (unbalanced)
\STATE Initialize dual variables $(\mathbf{u}, \mathbf{v})$
\STATE $i \gets 0$
\WHILE{$i < \text{max\_iter}$}
    \STATE $\mathbf{P} \gets \mathbf{K}$
    \STATE $\mathbf{u} \gets \mathbf{u} + \frac{\log(\mathbf{1}_{n \times I}) - \log(\text{bmm}(\mathbf{P}, \mathbf{v}[:, :, \mathrm{None}])[:, :, 0])}{\tau}$
    \STATE $\mathbf{P} \gets \mathbf{K}$
    \STATE $\mathbf{v} \gets \mathbf{v} + \frac{\log(\mathbf{1}_{n \times O}) - \log(\text{bmm}(\mathbf{u}[:, \mathrm{None}, :], \mathbf{P})[:, 0, :])}{\tau}$
    \STATE $i \gets i + 1$
\ENDWHILE

% Final transport plan
\STATE \textbf{Stage 2: Final transport plan}
\STATE $\mathbf{K} \gets \mathbf{u}[:, :, \mathrm{None}] \;\odot\; \mathbf{K} \;\odot\; \mathbf{v}[:, \mathrm{None}, :]$
\STATE \textbf{return} $\mathbf{K}$
\end{algorithmic}
\end{algorithm}

We define the transport plan \( T^{(H-I)} \) between H\&E images and weakly paired IHC images, capturing distributional differences such as texture structure, and the transport plan \( T^{I-{I'}} \) between generated and weakly paired IHC images, reflecting molecular expression imitation differences; we assume these two transport plans should be as similar as possible in the feature space. By minimizing the cost of the transport plan, regions with high self-information receive priority, facilitating the accurate representation of distributional relationships and semantic mappings. Specifically, given that H\&E images \( X \) and weakly paired IHC images \( Y \) share morphological and structural features, we define the minimal transport cost \( f_1 \) from \( x_{ij} \) to \( y_{ij} \); similarly, as weakly paired IHC \( Y \) and generated IHC \( Z \) share staining type and texture features, we define \( f_2 \) from \( y_{ij} \) to \( z_{ij} \). Consequently, an indirect transport plan 
$f_{\text{indirect}} = f_1 \cdot f_2$ can represent the sequential feature mapping through these two steps, while the direct transport plan 
$f_{\text{direct}}$ between H\&E and generated IHC images is computationally simpler due to their shared features and adherence to classical optimal transport assumptions.

  \begin{equation}
    \mathcal{L}_{tcyc}=\frac{1}{N^2}\sum_{i=1}^{N}\sum_{j=1}^{N}\left|\textbf{T}^{(H-I)}\cdot \textbf{T}^{(I-I')}-f_{direct}\right|
\end{equation}

where $f_{direct}$ represent $\textbf{T}^{(H-\hat{I})}$, $\hat{I}$ is virtual IHC image. Theoretically, by combining transport plans, the joint probability distribution \( P(H, \hat{I}) \) between H\&E images and generated IHC replaces the joint probability distributions \( P(H, I) \) between H\&E and weakly paired IHC, and \( P(I, I') \) between weakly paired IHC and generated IHC, thereby eliminating the influence of weakly paired data. The joint marginal distributions  can be formulated as:

\[
P(H, I') = \sum_{I} P(H, I) \cdot P(I, I')
\]

The all transport plans are computed based on the Unbalanced Optimal Transport (UOT) method. 

\subsection{Intra-batch Self-information Mining}
Optical density is an indirect indicator for assessing antibody concentration in IHC-stained images, and it serves as an important basis for clinically evaluating disease subtypes in IHC images. Optical density is highly correlated with abnormal regions in the image, where high optical density values, representing high antibody concentrations, often reflect abnormal biological expressions. Therefore, as a critical metric for evaluating pathological semantics, optical density can act as a self-information anchor to directly guide the generative model in focusing on key regions. To effectively model the pathological correlation among batch-level IHC images, first, we introduce optical density, which is proportional to the concentration of the stain~\cite{ref_24}. The amount of stain is the factor determining the OD at a wavelength as per the LambertBeerlaw~\cite{ref_46}. It can be formulated as:

\begin{equation}
    OD_C=-log_{10}(I_C/I_{0,C})=A*c_C
    \label{eq:od}
\end{equation}

where $I_0$,C and $I_C$ denote the light intensity entering and passing the specimen. We use traditional color deconvolution~\cite{ref_47}%
 for stain separation. Then we specifically select DAB stain’s OD values to generate the RGB image (IHC DAB). Due to the high optical density (OD) values in positive regions, which are typically much smaller in area compared to negative regions, the distribution becomes unbalanced. To amplify the influence of positive regions in self-information mining, we adjust Equation \ref{eq:od} to assign grayscale values to positive signals using a focal calibration map, referred to as Focal Optical Density (FOD).

\begin{equation}    
    O_C = (-log_{10}(I_C/I_{0,C}))^\alpha
\end{equation}
where O is the FOD with tunable focusing parameter $\alpha$ $>$ 1. We convert virtual IHC and real IHC to $O^F$ and $O^R$, which use ReLU to activate and flatten as vector $f^F_i$ and $f^R_i$. Then we calculate the optical density relational function between any two real IHC feature within the batch to obtain correlation matrix $\mathbf{M}$, which can be formulated as :

\[
\mathbf{M} =
\begin{bmatrix}
\text{cos}(\mathbf{f}_1, \mathbf{f}_1) & \text{cos}(\mathbf{f}_1, \mathbf{f}_2) & \cdots & \text{cos}(\mathbf{f}_1, \mathbf{f}_n) \\
\text{cos}(\mathbf{f}_2, \mathbf{f}_1) & \text{cos}(\mathbf{f}_2, \mathbf{f}_2) & \cdots & \text{cos}(\mathbf{f}_2, \mathbf{f}_n) \\
\vdots & \vdots & \ddots & \vdots \\
\text{cos}(\mathbf{f}_n, \mathbf{f}_1) & \text{cos}(\mathbf{f}_n, \mathbf{f}_2) & \cdots & \text{cos}(\mathbf{f}_n, \mathbf{f}_n)
\end{bmatrix}
\]

where $cos(\cdot,\cdot)$ denotes the Cosine similarity. $\mathbf{M}^R$ and $\mathbf{M}^F$ denotes real IHC and fake IHC optical density self-correlation map. Then, we propose $\mathcal{L}_{odc}$, which maximize the preservation of pathological correlation among batch samples. It can be formulated as:

\begin{equation}
    \mathcal{L}_{odc}=||M^R-M^F||+\frac{1}{N^2}||\sum_{i,j}O^R(i,j)-\sum_{i,j}O^F(i,j)||^2_2
\end{equation}

We computed the cosine similarity of multi-scale features between real IHC and fake IHC within the generator to construct the self-correlation matrices $M^{F'}$ and $M^{R'}$. These matrices act as a shorter gradient backpropagation path, thereby influencing the encoder. $\mathcal{L}_{\text{cc}}$, as a shorter gradient backpropagation path, influences the encoder.

\begin{equation}
    \mathcal{L}_{cc}=||M^{R'}-M^{F'}||
\end{equation}

\subsection{Loss functions}

The loss function term in the USI-GAN include adversial loss $\mathcal{L}_{adv}$, PatchNCE loss $\mathcal{L}_{patchnce}$, transport consistency loss $\mathcal{L}_{tcyc}$, correlation consistency loss $\mathcal{L}_{cc}$, integrated optical density correlation consistency loss $\mathcal{L}_{odc}$. Adversial loss can be formulated as:
\begin{equation}
    \mathcal{L}_{adv}=\mathbb{E}_{y\in Y}\mathrm{log} D(Y)+\mathbb{E}_{x\in X}\mathrm{log} (1-D(X))
\end{equation}

where $\mathbb{E}$ denotes the expectation, X and Y denote the source and target domain feature. G represent Generator and D represent Discriminator.  PatchNCE loss $\mathcal{L}_{nce}$ establishes cross-domain correlations by maximizing the mutual information between the input and output, which can be expressed as:

\begin{multline}
    \mathcal{L}_{nce}(\mathbf{v},\mathbf{v}^+,\mathbf{v}^-) = \\
    -\log\bigg[
        \frac{\mathrm{exp}(\mathbf{v} \cdot \mathbf{v}^+ / \tau)}
        {
        \mathrm{exp}(\mathbf{v} \cdot \mathbf{v}^+ / \tau) 
        + \sum_{n=1}^N \mathrm{exp}(\mathbf{v} \cdot \mathbf{v}^- / \tau)
        }
    \bigg]
\end{multline}
where $\mathbf{v}$, $\mathbf{v}^+$ and $\mathbf{v}^-$ are the embeddings of the anchor,positive and negative samples, respectively, $\tau$ is a temperature hyper-parameter. The PatchNCE can be formulated as:

\begin{equation}
    \mathcal{L}_{patchnce}(X)=\mathbb{E}_{x\in X}\sum^L_{l=1}\sum^{S_l}_{s=1}\mathcal{L}_{nce}(\mathrm{z},\mathrm{z}^+,\mathrm{z}^-)
\end{equation}

where L represent the layer seleted from the multi-layer of encoder feature. $S_l$ is the number of spatial location in each layer, $\mathrm{z}$ represent the anchor embedding from output image. the positive $\mathrm{z}^+$ is the embedding of the corresponding patch
from the input image, while the negatives $\mathrm{z}^-$ are embeddings of the non-corresponding ones.

The total loss for the USI-GAN is shown as follows:
\begin{equation}
\label{eq:loos}
\mathcal{L}_{total}=\lambda_{tcyc}\mathcal{L}_{tcyc}+\lambda_{cc}\mathcal{L}_{cc}+\mathcal{L}_{odc}+\mathcal{L}_{nce}+\mathcal{L}_{adv}
\end{equation}

\section{Experiments And Results}

\subsection{Datasets}
\subsubsection{MIST Dataset} 
In the MIST dataset~\cite{ref_7}, there are aligned H\&E-IHC patches for four different IHC stains (Ki67, ER, PR and HER2) critical to breast cancer diagnosis. All patches are of size 1,024 $\times$ 1,024 and non-overlapping.

\subsubsection{IHC4BC Dataset}

Similar to the MIST dataset, there are aligned H\&E-IHC patches for four different IHC stains (Ki67, PR, HER2, ER) in the IHC4BC dataset~\cite{ref_9}. The IHC4BC dataset contains more patches than the MIST dataset, containing approximately 90,000 H\&E-IHC pairs. The patch size of these pairs is 1,000 $\times$ 1,000.

\begin{table}[!h]
    \centering
    \label{dataset}
    \caption{Details of MIST and IHC4BC Public Benckmark Datasets. Paired Rate represent Pearson-R of H\&E and Weakly Paired IHC. }
    \resizebox{\columnwidth}{!}{
    \begin{tabular}{cccccc} % 所有列居中对齐
        \hline
        {Dataset} & {Staining Type} & {No.WSI} & {No.Pairs (train)} & {No.Pairs (test)} & Paired Rate  \\
        \hline
        \multirow{4}{*}{{MIST}} & Ki67 & 56 & 4,361 & 1,000 & 8.50\% \\
                                       & ER   & 56 & 4,153 & 1,000 & 11.82\% \\
                                       & PR   & 56 & 4,139 & 1,000 & 11.46\%\\
                                       & HER2 & 64 & 4,642 & 1,000 & 9.11\%\\
        % \hline
        % {BCI} & HER2 & 51 & 4,879 & 931 & 16.12\% \\
        \hline
        \multirow{4}{*}{{IHC4BC}} & Ki67 & 60 & 17,745 & 1,000 & 7.37\% \\
                                       & ER   & 59 & 26,395 & 1,000 & 16.62\% \\
                                       & PR   & 60 & 20,071 & 1,000 & 14.09\%\\
                                       & HER2 & 52 & 16,995 & 1,000 & 16.45\%\\
        \hline
    \end{tabular}}
\end{table}

{The H\&E-IHC image pairs in MIST and IHC4BC dataset are registered and achieve overall structual consistency. We employed the Pearson-R correlation coefficient to measure the grayscale correlation between pairs of H\&E and IHC images. Each patch was resized to $1024 \times 1024$ and cropped to $512 \times 512$ during training.}

\subsection{Experimental Settings}
Our method is implemented with Python based on PyTorch on a computer with Intel(R) Core(TM) i5-10400 CPU, 48 GB RAM, and a NVidia RTX A6000 GPU. All experiments were conducted with an image resolution of 1024×1024, and we trained our method with random 512×512 crops. We used 5-layer PatchGAN as the discriminator and ResNet-6Blocks as the generator for our method. For contrastive learning setting, we keep the same with CUT~\cite{ref_19}, e.g. 256 negative samples, temperature parameter $\tau = 0.07$ for PatchNCE loss, and batch size as 2. The USIGAN was trained for 80 epochs for MIST (about 2 days)  and 20 epochs for IHC4BC dataset (about 3 days). During training, we used the Adam optimizer with a linear decay scheduler and an initial learning rate of $2\times10^{-4}$. The hyperparameters in Eq.\ref{eq:loos} were set as: $\lambda_{tcyc}= 10,000$  and $\lambda_{cc}= 10$.

% All quantitative and qualitative experiment results reported in this study are based on the test set to evaluate the final performance of the proposed method.

\subsection{Evaluation Metrics}
{In this study, we employed three dimensions of metrics to evaluate our work. 
Pixel-level metrics include Structural Similarity Index (SSIM) and Peak Signal-to-Noise Ratio (PSNR). Pathology-related metrics consist of the IoD and Pearson correlation coefficient. Perceptual metrics include the Fréchet Inception Distance (FID)~\cite{ref_37}, Deep Image Structure and Texture Similarity (DISTS), and PHV$_{avg}$, which is described in~\cite{ref_18}. }

{For pathological evaluation, we utilized the Integrated Optical Density (IoD), following the clinical guidelines described in \cite{jensen2013quantitative}, to quantify the immunohistochemistry (IHC) images. Additionally, we calculated the Pearson correlation coefficient between the fluorescence intensities of real IHC datasets and the generated IHC images as a measure of pathological fidelity, as outlined in \cite{ref_24,ref_35}. Futhermore, we observed that cross-staining registration in pathological imaging often assumes that grayscale representations of images with different stains share similar distributions and content \cite{wodzinski2024regwsi}. 
Therefore, we computed the Pearson correlation coefficient of pixel values between the grayscale versions of the virtually stained results and the H\&E images, serving as a standard for content consistency. We use $R_{avg}$ represent virtual IHC clinical value.}

\begin{equation}
    R_{c} = \frac{\sum_{i=1}^{N} (I_{1}^{(i)} - \bar{I}_{1})(I_{2}^{(i)} - \bar{I}_{2})}{\sqrt{\sum_{i=1}^{N} (I_{1}^{(i)} - \bar{I}_{1})^2} \sqrt{\sum_{i=1}^{N} (I_{2}^{(i)} - \bar{I}_{2)^2}}}
\end{equation}

\begin{equation}
    R_{p} = \frac{
        \frac{1}{N-1} \sum_{i=1}^{N} (D_i - \bar{D})(O_i - \bar{O})
    }{
        \sqrt{
            \frac{1}{N-1} \sum_{i=1}^{N} (D_i - \bar{D})^2 \cdot 
            \frac{1}{N-1} \sum_{i=1}^{N} (O_i - \bar{O})^2
        }
    }
\end{equation}

\begin{equation}
    R_{avg} = (R_c+R_p)/2*100\%
\end{equation}

where $R_{c}$ and $R_{p}$ represents content preserving and pathological preserving Pearson correlation, respectively.  $N$ is total numbers of test dataset. $I_{1}^{(i)}$ and $ I_{2}^{(i)}$ are the grayscale values of the $i$-th pixel in each image. $\bar{I}_{1}$  and $\bar{I}_{2}$ are the mean grayscale values of the two images. $D_{i}$ and $O_{i}$ represents $i$-th image integrated density.

\begin{figure*}[!ht]
\centerline{\includegraphics[width=\textwidth]{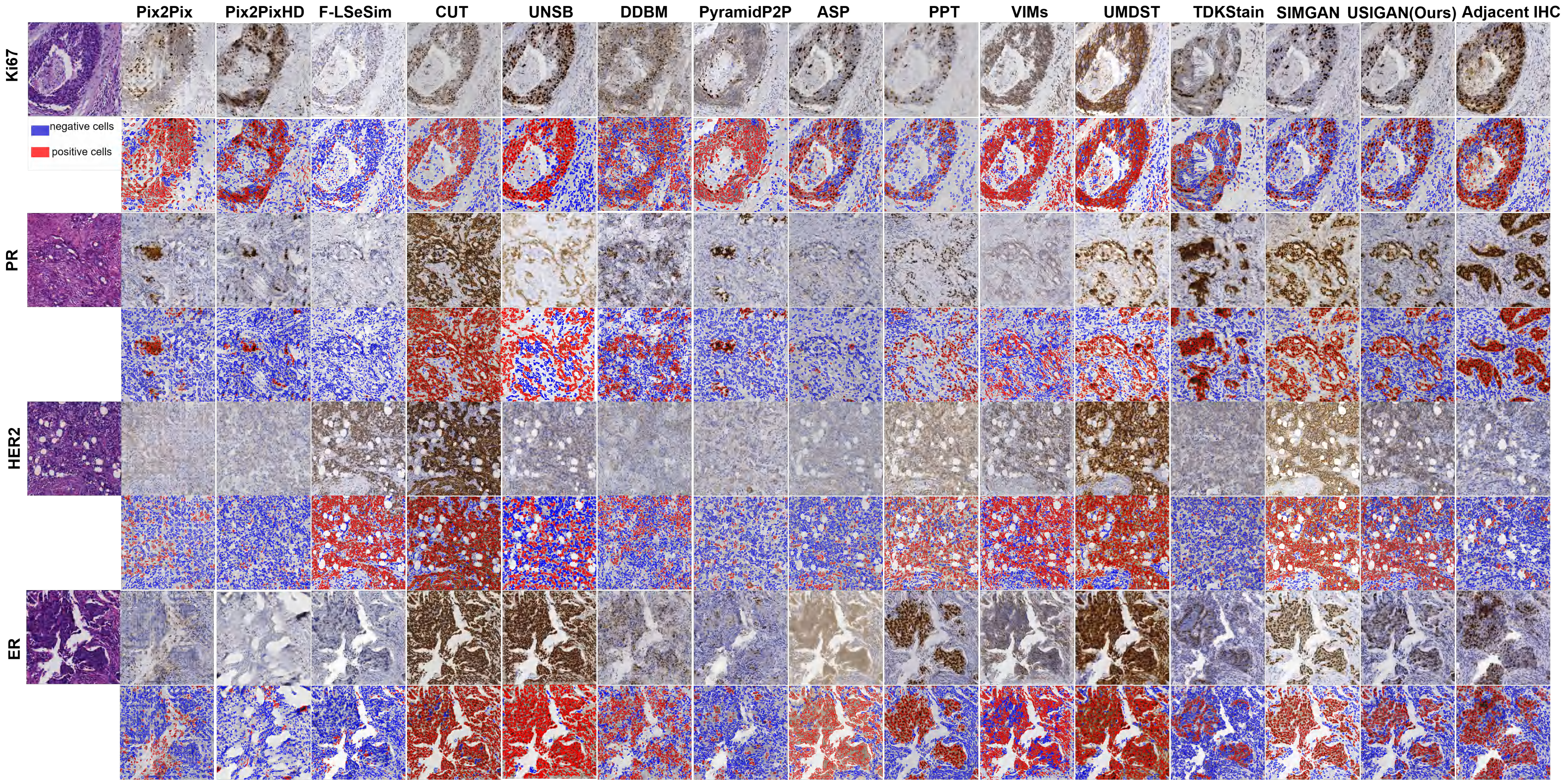}}
\caption{{Selected representative methods exhibit varying performances in virtual IHC staining results visualization on the MIST dataset. The quantitative comparison on different sate-of-art methods. Cell Segmentation and classification is performed using DeepLIIF~\cite{ref_53} as follow by \cite{pati2024accelerating}}}
\label{fig_e1_1}
\end{figure*}

\begin{figure*}[!ht]
\centerline{\includegraphics[width=\textwidth]{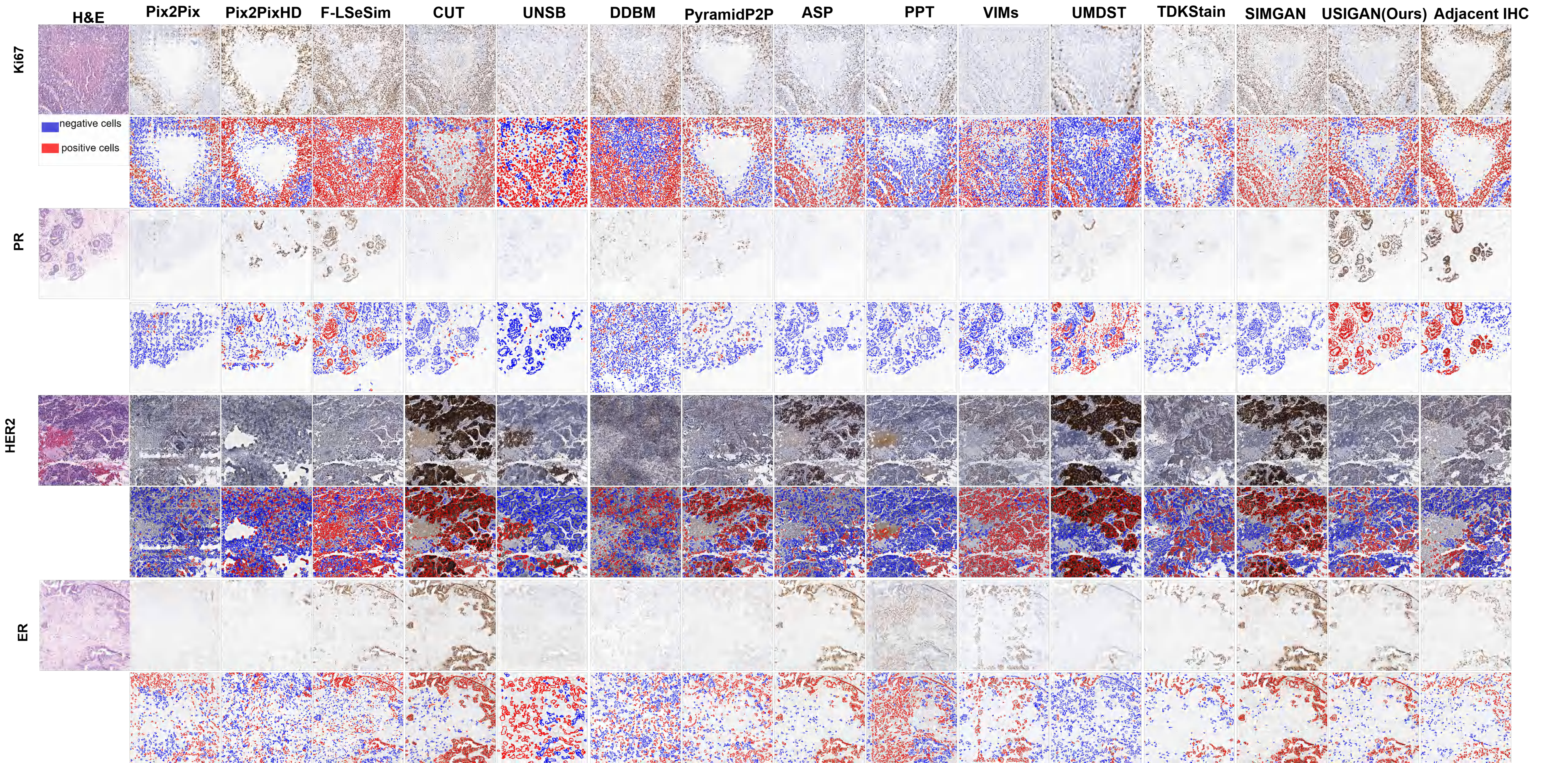}}
\caption{{Selected representative methods exhibit varying performances in virtual IHC staining results visualization on the IHC4BC dataset. The quantitative comparison on different sate-of-art methods. Cell Segmentation and classification is performed using DeepLIIF~\cite{ref_53} as follow by \cite{pati2024accelerating}}}
\label{fig_e2_2}
\end{figure*}

\subsection{Comparison with Competitive  Methods}

We compare our method with baselines and thirteen state-of-art Image-to-Image translation approches. It can be divided into the following 2 types: image translation SOTA and virtual staining SOTA methods.

% 1. \textbf{Image translation SOTA} based on building cross-domain correspondence by contrastive learning includes methods such as: Pix2Pix \cite{ref_14}, CycleGAN \cite{ref_28}, CUT \cite{ref_19}, QSAttn (\textit{CVPR'22}), MoNCE \cite{ref_26} (\textit{CVPR'22}), SRC \cite{ref_50} (\textit{CVPR'22}), LSeSim \cite{ref_52} (\textit{CVPR'21}), NEGCUT \cite{wang2021instance} (\textit{ICCV'21}), DECENT \cite{xie2022unsupervised} (\textit{NeurIPS'22}), SANTA \cite{xie2023unpaired} (\textit{CVPR'23}), ControlNet \cite{ref_controlnet} (\textit{CVPR'23}), DDBM \cite{ddbm} (\textit{ICLR'24}), UNSB \cite{ref_51} (\textit{ICLR'24}), EnCo \cite{cai2024rethinking} (\textit{AAAI'24}),StegoGAN \cite{wu2024stegogan} (\textit{CVPR'24}), and Pix2Pix-Turbo \cite{parmar2024one}.
% % 补了NEGCUG DECENT SANTA ControlNet,Pix2Pix-Turbo，EnCo,DDBM

% 2. \textbf{Virtual staining SOTA methods}, such as: PyramidPix2Pix \cite{ref_18} (\textit{CVPR'23}), ASP \cite{ref_7} (\textit{MICCAI'23}), UMDST \cite{umdst} (\textit{AAAI'22}), PPT \cite{ref_22} (\textit{MICCAI'24}), TDKStain \cite{ref_23} (\textit{MICCAI'24}), GramGAN \cite{gramgan} (\textit{TIP'24}), and SIMGAN \cite{ref_25} (\textit{TMI'25}).
% % UMDST GramGAN
%  PSPStain
% Note that, all results were obtained by their released code strictly.

\begin{table*}[ht]
  \centering
  \setlength\tabcolsep{5pt}
  \renewcommand{\arraystretch}{1.2}
  \caption{COMPARISON OF VIRTUAL STAINING PERFORMANCE ON \textbf{MIST} DATASET. THE BEST SCORES ARE IN {\color{red}\textbf{BOLD}} AND SECOND SCORES ARE IN {\color{blue}\underline{UNDERLINE}}, RESPECTIVELY. THE METRICA/METRICB INDICATES THE (BATCH SIZE SAME WITH USI-GAN/BATCH SIZE SET AS 1 FOR GAN-BASED METHOD) RESULTS.}
  \label{tab:mist_comparsion}
  
  \resizebox{\textwidth}{!}{
    \begin{tabular}{c|cc|cc|ccc|cc|cc|ccc}
      \toprule
      \multirow{3}{*}{Method} & 
      \multicolumn{7}{c|}{HER2} & \multicolumn{7}{c}{ER} \\
      \cmidrule(lr){2-15}
      & \multicolumn{2}{c|}{Quality (reference)} & \multicolumn{2}{c|}{Pathology} & \multicolumn{3}{c|}{Perception} & 
      \multicolumn{2}{c|}{Quality (reference)} & \multicolumn{2}{c|}{Pathology} & \multicolumn{3}{c}{Perception} \\
      \cmidrule(lr){2-15}
      & SSIM & PSNR & $R_{avg}\uparrow$ & IoD$^{\times10^7}$ & FID$\downarrow$ & DISTS$\downarrow$ & PHV$_{avg}\downarrow$ & 
      SSIM & PSNR & $R_{avg}\uparrow$ & IoD$^{\times10^7}$ & FID$\downarrow$ & DISTS$\downarrow$ & PHV$_{avg}\downarrow$ \\
      \midrule
       \multicolumn{15}{c}{\textbf{GAN-based Image-to-Image Translation SOTA}} \\
      \midrule
      Pix2Pix~\cite{ref_14} & 
      0.2022 / {\color{blue}\underline{0.2224 }}& 14.8224 / {\color{red}\textbf{15.244}}  & 37.44 / 40.29 & -5.6092 / -4.4196 & 107.40 / 73.32  & 0.2999 / 0.2912 & 0.5164 / 0.5260  & 
      0.2329 / {\color{red}\textbf{0.2535}} & 14.8243 /  {\color{red}\textbf{15.4573}} & 42.39 / 44.97 & -6.5789 / -5.646 & 87.29 /77.33 & 0.3393 / 0.3236 & 0.5565 /0.5889  \\
      % Pix2Pix$^\dag$~\cite{ref_14} & 
      % - & - & - & - & - & - & - & 
      % - & - & - & - & - & - & - \\

      % PyramidP2P$^\dag$~\cite{ref_18} & 
      % - & - & - & - & - & - & - & 
      % - & - & - & - & - & - & - \\
      
      {Pix2PixHD}~\cite{wang2018pix2pixHD} & 
      {\color{red}\textbf{0.2485}} / 0.2093 & 14.6117 / 14.5924 & 25.54 / 40.15 & -3.6936 / -3.8803 & 87.51 / 68.48 & 0.3156 / 0.2899 & 0.5665 / 0.4752 & 
      {\color{blue}\underline{0.2457}} / 0.2419 & 14.2915 / 14.7984 & 30.04 / 41.48 & -6.7141 / -5.3618 & 97.68 / 68.61 & 0.3630 / 0.3129 & 0.5643 / 0.5072 \\
      % Pix2PixHD$^\dag$ & 
      % - & - & - & - & - & - & - & 
      % - & - & - & - & - & - & - \\
      
      CycleGAN~\cite{ref_28} & 
      0.1805 / 0.2008 & 13.1507 / 13.6588  & 58.77 / 60.36 & -1.0736 / {\color{blue}\underline{-0.1638}} & 54.65 / 53.97 & 0.26681 / 0.2629 & 0.5053 / 0.5053 & 
      0.2191 / 0.2050 & 13.4436 / 13.1596 & 63.57 / 63.32 & -1.1561 / {\color{blue}\textbf{-0.8555}} & 37.44 / 39.99 & 0.2665 / 0.2655 & 0.4992 / 0.4976 \\

      % CycleGAN$^\dag$~\cite{ref_28} & 
      % - & - & - & - & - & - & - & 
      % - & - & - & - & - & - & - \\

     F-LseSim~\cite{ref_52} & 
      0.1813 / 0.1665 & 13.2261 / 12.5903 & 58.27 / 61/67 & -3.1648 / 1.1324 & 51.10 / 50.70  & 0.2685 / 0.2810 & 0.5335 / 0.5431 & 
      0.2132 / 0.1802 & 13.9336 / 13.2137 & 65.38 / 54.76 & -9.1023 / -4.4095 & 143.57 / 84.70 & 0.2903 / 0.2869 & 0.5451 / 0.5759 \\

      % F-LseSim$^\dag$~\cite{ref_52} & 
      % - & - & - & - & - & - & - & 
      % - & - & - & - & - & - & - \\

      CUT~\cite{ref_19} & 
      0.1453/ 0.1698 & 11.6753 / 12.4523 & 57.78 / 56.90 & 10.9532 / 5.435 & 69.55 / 52.55 & 0.3000 / 0.2845 & 0.5712 / 0.5405 & 
      0.1999 / 0.1954 & 13.2221 / 13.5516  & 63.44 / 60.52 & 2.1214 / 6.0702 & 59.95 / 79.75 & 0.2942 / 0.3006 & 0.5474 / 0.5674 \\

      % CUT$^\dag$~\cite{ref_19} & 
      % - & - & - & - & - & - & - & 
      % - & - & - & - & - & - & - \\

      MoNCE~\cite{ref_26}& 
      0.1525 / 0.164 & 11.9814 / 12.082 & 56.44 / 54.86 & 7.2754 / 6.4068 & 60.65 / 61.58 & 0.2959 / 0.2893 & 0.5652 / 0.5429 & 
      0.1886 / 0.1818 & 12.6758 / 12.6003 & 65.86 / 68.88 & 6.1803  / 2.7424 & 59.99 / 67.79 & 0.2870 / 0.2778 & 0.5430 / 0.5541 \\

      % MoNCE$^\dag$~\cite{ref_26}& 
      % - & - & - & - & - & - & - & 
      % - & - & - & - & - & - & - \\

      QSAttn~\cite{ref_40}& 
      0.1492 / 0.157 & 11.8613 / 11.8377 & 55.67 / 56.00 & 8.1264 / 9.0991 & 65.11 / 63.99 & 0.2925 / 0.2921 & 0.5632 / 0.5531 & 
      0.1808 / 0.2034 & 12.4868 / 13.026 & 65.79 / 58.89 & 6.944 / 5.0241 & 71.65 / 83.37 & 0.2957 / 0.3004 & 0.5701 / 0.5644 \\

      % QSAttn$^\dag$& 
      % - & - & - & - & - & - & - & 
      % - & - & - & - & - & - & - \\

      {NEGCUT}~\cite{wang2021instance}& 
      0.1534 / 0.1234 & 12.2013 / 10.6738 & 58.69 / 67.36 & 2.6799 / -2.3618 & 53.39 / 53.52 & 0.2956 / 0.2554 & 0.5181 / 0.5139 & 
      0.1874 / 0.1725 & 12.6041 / 11.9798 & 65.57 / 69.33 & 6.5302 / 4.7934 & 54.24 / 63.41 & 0.2830 / 0.3019 & 0.5482 / 0.5854 \\

      % NEGCUT$^\dag$& 
      % - & - & - & - & - & - & - & 
      % - & - & - & - & - & - & - \\

    SRC~\cite{ref_50}& 
      - / 0.1639 & - / 12.4392 & - / 44.46 & - / 0.9507 & - / 56.43 & - / 0.3053 & - / 0.5719 & 
       - / 0.1741 & - / 12.1321 & - / 66.87 & - / 5.3383 & - / 55.46 & - / 0.2930 & - / 0.5615 \\

      % SRC$^\dag$& 
      % - & - & - & - & - & - & - & 
      % - & - & - & - & - & - & - \\

      {DECENT}~\cite{xie2022unsupervised} &
      0.1719 / 0.1651 & 12.4824 / 12.4744 & 50.34 / 54.38 & 2.1423 / 4.6545 & 58.23 / 62.36  & 0.2987 / 0.2994 & 0.5673 / 0.5701 & 
      0.2044 / 0.1961 & 13.3857 / 13.076 & 65.97 / 61.81 & 2.7814 / 3.4981 & 68.77 / 61.81 & 0.2881 / 0.2863 & 0.5558 / 0.5546 \\

      % DECENT$^\dag$& 
      % - & - & - & - & - & - & - & 
      % - & - & - & - & - & - & - \\

      % SANTA~\cite{xie2023unpaired}& 
      % 0.1392 & 13.5393 & 57.29 & -10.1118 & 41.11 & 0.2718 & 0.5513 & 
      % 0.1658 & 13.4831 & 59.59 & -10.4834 & 35.70 & 0.2673 & 0.5588 \\

      % SANTA$^\dag$& 
      % - & - & - & - & - & - & - & 
      % - & - & - & - & - & - & - \\

      {EnCo}~\cite{cai2024rethinking}& 
      0.1604 / 0.1558 & 12.1925 / 12.1163 & 52.54 / 51.84 & 4.7079 / 7.279 & 84.43 / 72.53  & 0.2975 / 0.3050 & 0.5734 / 0.5736  & 
      0.1739 / 0.1788 & 12.6444 / 13.0233 & 59.72 / 63.51 & 1.4499 / 2.1424 & 67.58 / 61.33 & 0.2912 / 0.2856 & 0.5707 / 0.5567 \\

      % EnCo$^\dag$& 
      % - & - & - & - & - & - & - & 
      % - & - & - & - & - & - & - \\

    % StegoGAN~\cite{wu2024stegogan}& 
    %   - & - & - & - & - & - & - & 0.2076 & 12.7642 & 63.22 & 3.6211 & 44.52 & - & 0.5165  
    %    \\

      % StegoGAN$^\dag$& 
      % - & - & - & - & - & - & - & 
      % - & - & - & - & - & - & - \\
    \midrule
       \multicolumn{15}{c}{\textbf{Diffusion-based Image-to-Image Translation SOTA}} \\
      \midrule
    
      {UNSB}~\cite{ref_51}& 
      0.128 / - & 13.2194 / - & 58.36 / - & -1.0456 / - & 48.77 / - & 0.2695 / - & 0.5833 / - & 
      0.1572 / - & 13.3287 / - & 62.78 / - & -9.7522 / - & 34.49 / - & 0.2644 / - & 0.5603 / - \\

    {DDBM}~\cite{ddbm}& 
      0.1891 / - & 14.6195 / - & 14.87 / - & -7.3041 / - & 188.51 / - & 0.3442 / - & 0.5901 / - & 
      0.2094 / - & 14.5504 / - & 28.92 / - & -7.0279 / - & 180.32 / - & 0.3383 / - & 0.6001 / - \\

      {ControlNet}~\cite{ref_controlnet}& 
      0.1545 / - & 8.6484 / - & -1.74 / - & 9.6394 / - & 122.23 / - & 0.3846 / - & 0.6744 / - & 
      0.145 / - & 8.376 / - & 5.036 / - & 24.6374 / - & 154.64 / - & 0.3445 / - & 0.6351 / - \\
      {VIMs}\cite{dubey2024vimsvirtualimmunohistochemistrymultiplex}& 
      0.1595 / - & 13.3339 / - & 28.75 / - & -2.5329 / - & 116.25 / - & 0.3084 / - & 0.6257 / - & 
      0.191 / - & 13.7069 / - & 44.52 / - & -3.9112 / - & 111.54 / - & 0.3014 / - & 0.6144 / - \\
      
    \midrule
       \multicolumn{15}{c}{\textbf{ Virtual Staining SOTA}} \\
      \midrule

      PyramidP2P~\cite{ref_18} & 
      0.1944 / 0.208 & 14.5908 / 15.1779  & 38.60 / 41.73 & -5.8272 / -3.7689 & 103.07 / 78.89 & 0.2764 / 0.2814  & 0.5078 / 0.5185 & 
      0.2236 / 0.2346 & 14.7975 /  {\color{blue}\underline{15.2438}} & 43.73 / 46.54 & -5.5017 / -5.4274 & 91.71 / 74.46 & 0.3066 / 0.2989 & 0.5432 / 0.5158 \\
      
    PPT~\cite{ref_22}& 
      0.2008 / 0.2096 & 14.3922 / 14.6015 & 55.98 / 38.99 & -4.6784 / -3.4225 & 54.38 / 62.04 & 0.2884 / 0.3133 & 0.6194 / 0.5425 & 
      0.2086 / 0.1517 & 13.9607 / 12.5299 & 58.13 & -4.522 & 50.44 / 77.14 & 0.3124 / 0.2959 & 0.5311 / 0.5781 \\

      % PPT$^\dag$& 
      % - & - & - & - & - & - & - & 
      % - & - & - & - & - & - & - \\

      {UMDST}~\cite{umdst}& 
      - / 0.2185 & - / 12.6884 & - / 49.46 & - / {\color{red}\textbf{-0.0378}} & - / 70.41 & - / 0.3124 & - / 0.6045 & 
      - / 0.253 & - / 13.6299  & - / 63.22 & - / 3.6211 & - / 44.52 & - / 0.3123 & - / 0.5165 \\

      % UMDST$^\dag$& 
      % - & - & - & - & - & - & - & 
      % - & - & - & - & - & - & - \\

      % GramGAN~\cite{gramgan}& 
      % - & - & - & - & - & - & - & 
      % - & - & - & - & - & - & - \\

      % GramGAN$^\dag$& 
      % - & - & - & - & - & - & - & 
      % - & - & - & - & - & - & - \\

    ASP~\cite{ref_7}& 
      0.2025 / 0.1632 & {\color{blue}\underline{14.6295}} / 12.9524 & {\color{blue}\underline{76.31}} / 57.15   & -4.7957 / 0.91334  & 47.86 / 63.53 & 0.2792 / 0.2844  & 0.5020 /  0.5516 & 
      0.2143 / 0.1954 & 14.1445 / 13.5516 & 74.96 / 72.74  & -3.1462 / -2.3544 & 41.40 / 55.68  & 0.2646 / 0.2716 & 0.5071 / 0.5213 \\

      % ASP$^\dag$& 
      % - & - & - & - & - & - & - & 
      % - & - & - & - & - & - & - \\

      TDKStain~\cite{ref_7}& 
      0.1957 / 0.2025 & 14.5065 / 14.4854 & 44.41 / 45.71 & -1.8509 / -1.8614 & 58.64 / 65.99 & {\color{blue}\underline{0.2465}} / 0.2531 & {\color{blue}\underline{0.4718}} / 0.4759 & 
      0.213 / 0.2205 & 14.392 / 14.6813 & 49.66 / 45.88 & -2.6782 / -4.4095 & 48.93 / 73.01 & 0.2398 / 0.2860 & 0.4755 / 0.4995 \\

      % TDKStain$^\dag$& 
      % - & - & - & - & - & - & - & 
      % - & - & - & - & - & - & - \\

      SIMGAN~\cite{ref_25}& 
      0.1864 & 13.6776 & 73.73 & -1.4421 & {\color{blue}\underline{39.66}} & 0.2608 & 0.5093 & 
      0.2165 & 13.9746 & / {\color{blue}\underline{78.27}} & -1.4809 & {\color{blue}\underline{34.61}} & {\color{blue}\underline{0.2532}} & {\color{blue}\underline{0.4977}} \\

      \cellcolor{gray!20}USI-GAN(ours)& 
      \cellcolor{gray!20}0.1871 & \cellcolor{gray!20}14.0592 & \cellcolor{gray!20}{\color{red}\textbf{83.35}} & \cellcolor{gray!20}-1.6788 & \cellcolor{gray!20}{\color{red}\textbf{37.76}} & \cellcolor{gray!20}{\color{red}\textbf{0.2342}}  & \cellcolor{gray!20} {\color{red}\textbf{0.4708}}  & 
      \cellcolor{gray!20}0.2019 & \cellcolor{gray!20}13.7622 & \cellcolor{gray!20}{\color{red}\textbf{85.07}} & \cellcolor{gray!20}{\color{red}\textbf{-0.5109}} & \cellcolor{gray!20}{\color{red}\textbf{33.06}} & \cellcolor{gray!20}{\color{red}\textbf{0.2338}} & \cellcolor{gray!20}{\color{red}\textbf{0.4694}} \\
      
      \bottomrule

       \hline

       \multirow{3}{*}{Method} & 
      \multicolumn{7}{c|}{Ki67} & \multicolumn{7}{c}{PR} \\
      \cmidrule(lr){2-15}
      & \multicolumn{2}{c|}{Quality (reference)} & \multicolumn{2}{c|}{Pathology} & \multicolumn{3}{c|}{Perception} & 
      \multicolumn{2}{c|}{Quality (reference)} & \multicolumn{2}{c|}{Pathology} & \multicolumn{3}{c}{Perception} \\
      \cmidrule(lr){2-15}
      & SSIM & PSNR & $R_{avg}\uparrow$ & IoD$^{\times10^7}$ & FID$\downarrow$ & DISTS$\downarrow$ & PHV$_{avg}\downarrow$ & 
      SSIM & PSNR & $R_{avg}\uparrow$ & IoD$^{\times10^7}$ & FID$\downarrow$ & DISTS$\downarrow$ & PHV$_{avg}\downarrow$ \\
      \midrule
       \multicolumn{15}{c}{\textbf{ GAN-based Image-to-Image Translation SOTA}} \\
      \midrule
      Pix2Pix~\cite{ref_14} & 
      0.2397 / {\color{blue}\underline{0.2766}}  & 14.6725 / {\color{red}\textbf{15.8554}} & 37.14 / 41.02 & -2.9183 / -3.5921 & 97.35 / 87.33 & 0.2667 / 0.3296 & 0.4885 / 0.5693 & 
      0.2321 / {\color{blue}\underline{0.2936}} & 14.8758 / {\color{red}\textbf{15.8001}} & 43.27 / 48.29 & -6.5099 / 0.8295 & 81.88 / 93.51 & 0.3016 / 0.3534 & 0.5181 / 0.6001 \\
      % Pix2Pix$^\dag$~\cite{ref_14} & 
      % - & - & - & - & - & - & - & 
      % - & - & - & - & - & - & - \\

      % PyramidP2P$^\dag$~\cite{ref_18} & 
      % - & - & - & - & - & - & - & 
      % - & - & - & - & - & - & - \\
      
      {Pix2PixHD}~\cite{wang2018pix2pixHD} & 
      0.2483 / 0.2435 &  14.8874/ 14.4045 & 23.51 / 38.70 & -6.2699 / -0.5209 & 369.76 / 65.69 & 0.4193 / 0.2729 &  0.6437 / 0.4799 & 
      {\color{red}\textbf{0.3658}} / 0.2763 & 13.3915 / {\color{blue}\underline{15.0074}}  & 40.26 / 41.90 & -6.5009 / -5.7457 & 197.80 / 91.64 & 0.5374 / 0.3420 & 0.7480 / 0.5600 \\
      % Pix2PixHD$^\dag$ & 
      % - & - & - & - & - & - & - & 
      % - & - & - & - & - & - & - \\
      
      CycleGAN~\cite{ref_28} & 
      0.2294 / 0.2317 & 13.9732 / 14.0034 & 58.08 / 57.70 & -0.41 / -0.9368 & 33.97 / 33.54 & 0.2433 / 0.2444 & 0.4747 / 0.4786 & 
      0.2132 / 0.2139 & 13.3125 / 13.3928 & 66.23 / 64.00 & {\color{blue}\underline{-0.4074}} / -0.4639 & 40.21 / 40.38 & 0.2656 / 0.2666 & 0.4945 / 0.4976 \\

      % CycleGAN$^\dag$~\cite{ref_28} & 
      % - & - & - & - & - & - & - & 
      % - & - & - & - & - & - & - \\

     F-LseSim~\cite{ref_52} & 
      0.2349 / 0.191 & 13.9824 / 12.7051  & 48.06 / 57.51   & -3.8521 / 7.3051 & 66.63 / 67.47 & 0.2643 / 0.2810 & 0.5553 / 0.5399 & 
      0.1961 / 0.1848 & 13.174 / 13.344 & 63.74 / 70.69 & -10.355 / -3.6976 & 103.67 / 58.28 & 0.2910 / 0.2827 & 0.5726 / 0.5551 \\

      % F-LseSim$^\dag$~\cite{ref_52} & 
      % - & - & - & - & - & - & - & 
      % - & - & - & - & - & - & - \\

      CUT~\cite{ref_19} & 
      0.2174 / 0.2011 & 13.7533 / 12.7791  & 64.87 /65.47 & 3.9548 / 7.7662 & 54.42 / 59.37 & 0.2665 / 0.2756 & 0.5267 / 0.5418 & 
      0.205 / 0.2012  & 12.9842 / 12.7852 & 57.38 / 53.53 & 4.2059 / 4.2655 & 59.05 / 60.76 & 0.2966 / 0.2958  & 0.5585 / 0.5582 \\

      % CUT$^\dag$~\cite{ref_19} & 
      % - & - & - & - & - & - & - & 
      % - & - & - & - & - & - & - \\

      MoNCE~\cite{ref_26}& 
      0.2188/ 0.2158 & 13.3704 / 13.1576  & 66.26 / 61.89 & 4.2888 / 4.1535 & 55.54 / 56.72 & 0.2669 / 0.2696 & 0.5446 / 0.5467 & 
      0.2366 / 0.2216 & 14.2132 / 13.5972 & 69.28 / 67.89 & -1.0985 / 1.9259 & 48.56 / 42.50 & 0.2806 / 0.2778 & 0.5322 / 0.5271 \\

      % MoNCE$^\dag$~\cite{ref_26}& 
      % - & - & - & - & - & - & - & 
      % - & - & - & - & - & - & - \\

      QSAttn~\cite{ref_40}& 
      0.1985 / 0.2122 & 13.437 / 13.4907 & 60.18 / 66.02 & 3.4735 / 4.4797 & 49.45 / 58.86 & 0.2712 / 0.2749 & 0.5468 / 0.5413 & 
      0.1984 / 0.2000 & 12.6366 / 12.3894 & 60.09 / 59.93 & 7.0856 / 7.5200 & 69.78 / 73.58 & 0.2968 / 0.3054 & 0.5633 / 0.5684 \\

      % QSAttn$^\dag$& 
      % - & - & - & - & - & - & - & 
      % - & - & - & - & - & - & - \\

      {NEGCUT}~\cite{wang2021instance}& 
      0.1924 / 0.2158 & 12.4626 / 13.2313 & 45.37 / 63.51 & 7.0443 / 5.1103 & 67.22 / 59.17 & 0.2672 / 0.2801 & 0.5399 / 0.5650 & 
      0.2086 / 0.2065 & 13.2525 / 12.9201 & 69.26 / 67.38 & 1.5486 / 3.4305 & 53.659 / 54.48 & 0.2847 / 0.2889 & 0.5514 / 0.5574 \\

      % NEGCUT$^\dag$& 
      % - & - & - & - & - & - & - & 
      % - & - & - & - & - & - & - \\

    SRC~\cite{ref_50}& 
      - / 0.2286 & - / 13.7734 & - / 63.60 & - / 1.582 & - / 61.81 & - / 0.2671 & - / 0.5359  & 
      - / 0.1984 & - / 12.6936 & - / 72.75 & - / 3.4875 & - / 60.31  & - / 0.2814 & - / 0.5449 \\

      % SRC$^\dag$& 
      % - & - & - & - & - & - & - & 
      % - & - & - & - & - & - & - \\

      {DECENT}~\cite{xie2022unsupervised}& 
      0.2368 / 0.2223 & 14.1949 / 14.1141 & 62.47 / 62.61 & 1.1197 / 0.9440 & 54.27 / 60.58 & 0.2646 / 0.2708 & 0.5325 / -0.5473 & 
      0.2072 / 0.2098 & 13.3525 / 13.462 & 68.75 / 66.16 & 1.7774 / 0.8075 & 56.39 / 61.50 & 0.2821 / 0.2905 & 0.5542 / 0.5585 \\

      % DECENT$^\dag$& 
      % - & - & - & - & - & - & - & 
      % - & - & - & - & - & - & - \\

      % SANTA~\cite{xie2023unpaired}& 
      % 0.1775 & 14.1715 & 57.93 & -7.804 & 28.19 & 0.2342 & 0.5169 & 
      % 0.1679 & 13.7335 & 60.80 & -10.9632 & {\color{red}\textbf{33.81}} & 0.2648 & 0.5516 \\

      % SANTA$^\dag$& 
      % - & - & - & - & - & - & - & 
      % - & - & - & - & - & - & - \\

      {EnCo}~\cite{cai2024rethinking}& 
      0.1954 /0.2302 & 13.871 / 14.0677 & 60.09 / 59.62 & 1.106 / 1.5699 & 55.56 / 67.47 & 0.2654 / 0.2602 & 0.5472 / 0.5399 & 
      0.1871 / 0.1927  & 12.471 / 12.7945 & 65.71 / 64.66 & 3.3781 / 3.2650 & 81.47 / 77.99 & 0.2900 / 0.2924 & 0.5574 / 0.5646 \\

      % EnCo$^\dag$& 
      % - & - & - & - & - & - & - & 
      % - & - & - & - & - & - & - \\

    % StegoGAN~\cite{wu2024stegogan}& 
    %   - & - & - & - & - & - & - & 
    %   0.2002 & 13.0635 & - & - & 44.08 & - & 0.5002 \\

     \midrule
       \multicolumn{15}{c}{\textbf{ Diffusion-based Image-to-Image Translation SOTA}} \\
      \midrule
    
      {UNSB}~\cite{ref_51}& 
      0.1765 / - & 14.052 / - & 60.30 / - & -8.0121 / - & 39.29 / - & 0.2527 / - & 0.5684 / - & 
      0.1659 / - & 13.6937 / - & 61.68 / - & -10.8229 / - & 34.90 / - & 0.2793 / - & 0.5847 \\

    {DDBM}~\cite{ddbm}& 
      0.2085 / - & 14.0552 / - & 34.81 / - & 2.2744 / - & 150.88 / - & 0.2949 / - & 0.5773 / - & 
      0.2025 / - & 13.3431 / - & 41.17 / - & -6.6537 / - & 163.58 / - & 0.3094 / - & 0.5879 \\

      {ControlNet}~\cite{ref_controlnet}& 
      0.2744 / - & 10.2626 / - & 13.48 / - & 16.4868 / - & 282.97 / - & 0.3831 / - & 0.6683 / - & 
      0.1382 / - & 9.0184 / - & 5.944 / - & 8.3654 / - & 134.27 / - & 0.3670 / - & 0.6537 / - \\
      {VIMs}\cite{dubey2024vimsvirtualimmunohistochemistrymultiplex}& 
      0.1975 / - & 14.1509 / -  & 34.66 / - & -1.354 / - & 97.20 / - & 0.2986 / - & 0.6107 / - & 
      0.1941 / - & 13.9057 / - & 47.37 / - & -4.3369 / - & 129.18 / - & 0.3090 / - & 0.6188 / - \\
    \midrule
       \multicolumn{15}{c}{\textbf{ Virtual Staining SOTA}} \\
      \midrule

      PyramidP2P~\cite{ref_18} & 
      0.2823 / {\color{red}\textbf{0.2846}}  & 15.2563 / {\color{blue}\underline{15.7009}} & 35.47 / 41.47 & -2.2201 / -2.9104 & 111.51 / 78.72 & 0.3203 / 0.3017 & 0.5766 / 0.5494 & 
      0.2540 / 0.2455 & 15.2238 / 15.4207 & 46.38 / 49.32 & -5.5563 / -4.5925 & 92.63 / 76.69 & 0.3053 / 0.3042 & 0.5465 / 0.5432 \\
      
    PPT~\cite{ref_22}& 
      0.2663 / 0.2516 & 14.9242 / 15.5841 & 51.97 / 52.35 & -2.5072 / -3.8480 & 48.64 / 119.88 & 0.2788 / 0.3391 & 0.5241 / 0.6085 & 
      0.2256 / 0.182 & 14.3989 / 13.3685 & 63.20 / 55.76 & -6.2238 / -1.6585 & 50.31 / 65.43 & 0.2932 / 0.2966 & 0.5232 / 0.5595 \\

      % PPT$^\dag$& 
      % - & - & - & - & - & - & - & 
      % - & - & - & - & - & - & - \\

      \textcolor{red}{UMDST}~\cite{umdst}& 
      - / 0.2256 & - / 12.3384  & - / 18.94 & - / -0.787 & - / 64.31 & - / 0.3562  & - / 0.6478 & 
      - / 0.2518 & - / 13.6979 & - / 67.73 & - / -2.5224 & - / 58.40 & - /  0.3040  & - / 0.5987 \\

      % UMDST$^\dag$& 
      % - & - & - & - & - & - & - & 
      % - & - & - & - & - & - & - \\

      % GramGAN~\cite{gramgan}& 
      % - & - & - & - & - & - & - & 
      % - & - & - & - & - & - & - \\

      % GramGAN$^\dag$& 
      % - & - & - & - & - & - & - & 
      % - & - & - & - & - & - & - \\

    ASP~\cite{ref_7}& 
      0.2163 / 0.2123 & 14.525 / 14.1183 & 71.27 / 66.29 & -1.8027 / 0.3118 & 44.89 / 74.11 & 0.2464 / 0.2561 & 0.4922 / 0.5268 & 
      0.2178 / 0.1848 & 14.3087 / 13.344 & 74.34 / 70.69 & -5.19 / -3.6976 & 41.59 / 58.28 & 0.2554 / 0.2827 & 0.4924 / 0.5551 \\

      % ASP$^\dag$& 
      % - & - & - & - & - & - & - & 
      % - & - & - & - & - & - & - \\

      TDKStain~\cite{ref_23}& 
      0.2422 / 0.2525 & 15.0263 / 15.0208 & 42.58 / 42.26 & {\color{red}\textbf{-0.1888}} / -2.7421 & 61.98 / 64.80 & {\color{blue}\underline{0.2398}} / 0.2638 & {\color{blue}\underline{0.4892}} / 0.4900 & 
      0.2313 / 0.2279 & 14.7935 / 14.6793 & 48.38 / 47.76 & -3.2448 / -3.8214 & 57.90 / 64.18 & 0.2601 / 0.2656 & 0.4929 / {\color{blue}\underline{0.4873}} \\

      % TDKStain$^\dag$& 
      % - & - & - & - & - & - & - & 
      % - & - & - & - & - & - & - \\

      SIMGAN~\cite{ref_25} & 
      0.2322 & 14.2495 & {\color{blue}\underline{73.17}} & -0.6221 & {\color{blue}\underline{28.52}} & 0.2483 & 0.5022 & 
      0.2071 & 13.8286 & {\color{blue}\underline{85.16}} & {\color{red}\textbf{-0.2687}} & {\color{blue}\underline{35.87}} & {\color{blue}\underline{0.2521}} & 0.4995 \\

      \cellcolor{gray!20}USI-GAN(ours)& 
      \cellcolor{gray!20}0.2317 & \cellcolor{gray!20}14.4027 & \cellcolor{gray!20}{\color{red}\textbf{77.75}} & \cellcolor{gray!20}{\color{blue}\underline{-0.2680}} & \cellcolor{gray!20}{\color{red}\textbf{27.36}} & \cellcolor{gray!20}{\color{red}\textbf{0.2351}} &\cellcolor{gray!20} {\color{red}\textbf{0.4702}} & 
      \cellcolor{gray!20}0.2163 & \cellcolor{gray!20}14.2176 & \cellcolor{gray!20}{\color{red}\textbf{87.20}} & \cellcolor{gray!20}-2.0699 & \cellcolor{gray!20}{\color{red}\textbf{34.64}} & \cellcolor{gray!20}{\color{red}\textbf{0.2339}} & \cellcolor{gray!20}{\color{red}\textbf{0.4560}} \\
      \bottomrule
    \end{tabular}
  }
\end{table*}

\begin{table*}[ht]
  \centering
  \setlength\tabcolsep{5pt}
  \renewcommand{\arraystretch}{1.2}
  \caption{COMPARISON OF VIRTUAL STAINING PERFORMANCE ON \textbf{IHC4BC} DATASET. THE BEST SCORES ARE IN {\color{red}\textbf{BOLD}} AND SECOND SCORES ARE IN {\color{blue}\underline{UNDERLINE}}, RESPECTIVELY. THE METRICA/METRICB INDICATES THE (BATCH SIZE SAME WITH USI-GAN/BATCH SIZE SET AS 1 FOR GAN-BASED METHOD) RESULTS.}
  \label{tab:ihc4bc_comparsion}
  
  \resizebox{\textwidth}{!}{
    \begin{tabular}{c|cc|cc|ccc|cc|cc|ccc}
      \toprule
      \multirow{3}{*}{Method} & 
      \multicolumn{7}{c|}{HER2} & \multicolumn{7}{c}{ER} \\
      \cmidrule(lr){2-15}
      & \multicolumn{2}{c|}{Quality (reference)} & \multicolumn{2}{c|}{Pathology} & \multicolumn{3}{c|}{Perception} & 
      \multicolumn{2}{c|}{Quality (reference)} & \multicolumn{2}{c|}{Pathology} & \multicolumn{3}{c}{Perception} \\
      \cmidrule(lr){2-15}
      & SSIM & PSNR & $R_{avg}\uparrow$ & IoD$^{\times10^7}$ & FID$\downarrow$ & DISTS$\downarrow$ & PHV$_{avg}\downarrow$ & 
      SSIM & PSNR & $R_{avg}\uparrow$ & IoD$^{\times10^7}$$\downarrow$ & FID$\downarrow$ & DISTS$\downarrow$ & PHV$_{avg}\downarrow$ \\
      \midrule
       \multicolumn{15}{c}{\textbf{GAN-based Image-to-Image Translation SOTA}} \\
      \midrule
      
      Pix2Pix~\cite{ref_14} & 
      0.1248 / {\color{red}\textbf{0.1619}} & 11.37 / {\color{red}\textbf{12.5642}} & 47.68 / 46.80 & -7.1422 / -9.711 & 93.70 / 173.59 & 0.2855 / 0.4515 & 0.5242 / 0.6890 & 
      0.4628 / 0.4695 & {\color{blue}\underline{20.5881}} / 20.3643 & 44.96 / 39.67 & -1.9269 / -2.0841 & 102.54 / 149.93 & 0.2924 / 0.2938 & 0.4537 / 0.4744 \\
      % Pix2Pix$^\dag$~\cite{ref_14} & 
      % - & - & - & - & - & - & - & 
      % - & - & - & - & - & - & - \\

      % PyramidP2P$^\dag$~\cite{ref_18} & 
      % - & - & - & - & - & - & - & 
      % - & - & - & - & - & - & - \\
      
      {Pix2PixHD}~\cite{wang2018pix2pixHD} & 
        0.143 / {\color{blue}\underline{0.1473}} & 11.2992 / 10.9458 & 42.34 / 38.46 & -5.9182 / -10.4474 & 108.60 / 148.99 & 0.3754 / 0.3667 & 0.5946 / 0.6052 & 
      {\color{blue}\underline{0.502}} / 0.4757 & {\color{red}\textbf{20.8655}} / 20.2985 & 37.87 / 41.67 & -2.1063 / -1.0089 & 76.88 / 44.79 & 0.3384 / 0.2428 & 0.5061 / 0.4265 \\
      % Pix2PixHD$^\dag$ & 
      % - & - & - & - & - & - & - & 
      % - & - & - & - & - & - & - \\
      
      CycleGAN~\cite{ref_28} & 
      0.128 / 0.1269 & 10.9497 / 10.8503 & 59.13 / 61.80 & -9.5066 / -7.5383 & 53.60 / {\color{blue}\underline{49.80}} & {\color{blue}\underline{0.2452}} / 0.2629 & {\color{blue}\underline{0.4687}} / 0.4716 & 
      0.484 / 0.4615 & 19.6816 / 18.5512 & 69.39 / 73.18 & -0.4863 / 0.9928 & 39.04 / 39.94 & 0.2276 / 0.2654 & 0.3859 /0.3888  \\

      % CycleGAN$^\dag$~\cite{ref_28} & 
      % - & - & - & - & - & - & - & 
      % - & - & - & - & - & - & - \\

     F-LseSim~\cite{ref_52} & 
      0.1275 / 0.1243 & 10.8599 / 10.7016 & 54.81 / 58.56 & -13.4688 / {\color{blue}\underline{0.2620}} & 90.63 / 71.41 & 0.2647 / 0.2568 & 0.5324 / 0.5237 & 
      0.4391 / 0.4175 & 18.3421 / 18.1839 & 73.00 / 76.88 & 0.1354 / 1.0490 & 61.17 / 56.97 & 0.2667 / 0.2739 & 0.4683 / 0.4870 \\

      % F-LseSim$^\dag$~\cite{ref_52} & 
      % - & - & - & - & - & - & - & 
      % - & - & - & - & - & - & - \\

      CUT~\cite{ref_19} & 
      0.1214 / 0.1195 & 10.6364 / 10.338 & 66.00 / 63.32 & -2.7794 / 3.0142 & 53.14 / 51.73 & 0.2579 / 0.2629 & 0.5186 / 0.5364 & 
      0.413 / 0.4396 & 17.9916 / 18.6876 & 73.63 / 64.06 & 1.7072 / 0.2348 & 39.27 / 38.29 & 0.2442 / 0.2763 & 0.4589 / 0.4575 \\

      % CUT$^\dag$~\cite{ref_19} & 
      % - & - & - & - & - & - & - & 
      % - & - & - & - & - & - & - \\

      MoNCE~\cite{ref_26}& 
      0.1175 / 0.1211 & 10.3727 / 10.6194 & 64.73 / 68.16 & -1.3287 / -1.7506 & 60.06 / 53.57 & 0.2556 / 0.2529 & 0.5167 / 0.5160 & 
      0.4286 / 0.4242 & 17.9131 / 18.2314 & 75.81 / 68.88 & 1.6675 / 1.7383  & 46.64 / 44.81 & 0.2619 / 0.2622 & 0.4612 / 0.4695 \\

      % MoNCE$^\dag$~\cite{ref_26}& 
      % - & - & - & - & - & - & - & 
      % - & - & - & - & - & - & - \\

      QSAttn~\cite{ref_40}& 
      0.1228 / 0.1194 & 10.8112 / 10.6096 & 64.11 / 63.19 & -2.5190 / -2.0231 & 55.65 / 57.57 & 0.2584 / 0.2659 & 0.5182 / 0.5299 & 
      0.442 / 0.4644 & 18.5162 / 18.7556 & 63.48 / 67.34 & 0.1457 / 0.4662 & 39.35 / 40.83 & 0.2800 / 0.2659  & 0.4637 / 0.4472 \\

      % QSAttn$^\dag$& 
      % - & - & - & - & - & - & - & 
      % - & - & - & - & - & - & - \\

      {NEGCUT}~\cite{wang2021instance}& 
      0.1204 / 0.1234 & 10.603 / 10.6738 & 62.32 / 68.16 & -3.2282 / -2.3618 & 57.79 / 55.51 & 0.2648 / 0.2554 & 0.5288 / 0.5139 & 
      0.4056 / 0.4042 & 17.5892 / 16.8981 & 73.37 / 68.55 & 1.6634 / 2.8059 & 48.91 / 54.91 & 0.2700 / 0.2842 & 0.4986 / 0.5023 \\

      % NEGCUT$^\dag$& 
      % - & - & - & - & - & - & - & 
      % - & - & - & - & - & - & - \\

    SRC~\cite{ref_50}& 
     - / 0.1229 & - / 10.7725 & - / 63.87 & - / -2.6529 & - / 54.74 & - / 0.2630 & - / 0.5318 & 
      - / 0.4341 &  - / 18.3454 & - / 74.09 & - / 2.0176 & - / 40.91 & - / 0.2568 & - / 0.4595 \\

      % SRC$^\dag$& 
      % - & - & - & - & - & - & - & 
      % - & - & - & - & - & - & - \\

      DECENT~\cite{xie2022unsupervised}& 
      0.1202 / 0.1246  & 10.9916 / 10.8228 & 63.99 / 66.66 & -6.319 / -2.6494  & 57.66 / 56.55  & 0.2497 / 0.2704  & 0.4986 / 0.5407  & 
      0.4227 / 0.4063 & 18.211 / 17.7322 & 71.16 / 72.69 & -0.8704 / 1.6861 & 46.79 / 51.41 & 0.2651 / 0.2728 & 0.4711 / 0.4966 \\

      % DECENT$^\dag$& 
      % - & - & - & - & - & - & - & 
      % - & - & - & - & - & - & - \\

      % SANTA~\cite{xie2023unpaired}& 
      % 0.1015 & 10.6196 & 66.34 & -16.6829 & 52.04 & 0.2375 & 0.5031 & 
      % 0.4116 & 19.2854 & 67.63 & -3.0357 & {\color{red}\textbf{31.25}} & {\color{blue}\underline{0.2192}} & 0.4215 \\

      % SANTA$^\dag$& 
      % - & - & - & - & - & - & - & 
      % - & - & - & - & - & - & - \\

      {EnCo}~\cite{cai2024rethinking}& 
      0.1173 / 0.119 & 10.4794 / 10.7763 & 63.76 / 65.19  & -6.299 / -4.9355  & 71.14 / 59.60  & 0.2770 / 0.2689  & 0.5540 / 0.5556 & 
      0.4063 / 0.433 & 18.2537 / 18.5431 & 65.07 / 69.80 & 0.2648 / 0.2673 & 57.16 / 43.80 & 0.2719 / 0.2673 & 0.4946 / 0.4758 \\

      % EnCo$^\dag$& 
      % - & - & - & - & - & - & - & 
      % - & - & - & - & - & - & - \\

    % StegoGAN~\cite{wu2024stegogan}& 
    %   - & - & - & - & - & - & - & 
    %   - & - & - & - & - & - & - \\

      % StegoGAN$^\dag$& 
      % - & - & - & - & - & - & - & 
      % - & - & - & - & - & - & - \\
    \midrule
       \multicolumn{15}{c}{\textbf{Diffusion-based Image-to-Image Translation SOTA}} \\
      \midrule
    
      UNSB~\cite{ref_51}& 
      0.1087 / - & 11.1355 / - & 56.99 / - & -17.3059 / - & 49.04 / - & 0.2568 / - & 0.5809 / - & 
      0.4020 / - & 19.4709 / - & 60.03 / - & -3.2319 / - & 55.30 / - & 0.2617 / - & 0.5003 / - \\

    {DDBM}~\cite{ddbm}& 
      0.1067 / - & {\color{blue}\underline{11.6096}} / - & 31.14 / - & -11.335 / - & 198.794 / - & 0.3950 / - & 0.6627 / - & 
      0.4529 / - & 20.3453 / -  & 22.93 / - & -3.0707 / - & 279.76 / - & 0.3612 / - & 0.5676 / - \\

      {ControlNet}~\cite{ref_controlnet}& 
      0.0983 / - & 7.8967 / - & 23.63 / - & 15.6894 / - & 140.56 / - & 0.3718 / - & 0.6902 / - & 
      0.2639 / - & 7.4072 / - & 13.03 / - & 39.5472 / - & 232.41 / - & 0.4038 / - & 0.7062 / - \\
      {VIMs}\cite{dubey2024vimsvirtualimmunohistochemistrymultiplex}& 
      0.1272 / - & 11.4642 / - & 40.34 / - & -12.5726 / - & 81.40 / - & 0.3084 / -  & 0.6164 / - & 
      0.4239 / - & 17.7305 / - & 45.63 / - & -1.4041 / - & 89.95 / - & 0.3014 / - & 0.5134 \\
      
  \midrule
       \multicolumn{15}{c}{\textbf{Virtual Staining SOTA}} \\
      \midrule

    PyramidP2P~\cite{ref_18} & 
      0.1162 / 0.1297 & 11.1824 / 11.5591 & 45.24 / 47.26 & -8.0986 / -7.8924  & 92.52 / 74.99 & 0.2776 / 0.2869 & 0.5095 / 0.5204 & 
      0.4603 / 0.4462  & 20.675 / 20.6914  & 48.78 / 47.93 & -1.6351 / -1.6706  & 73.05 / 63.48 & 0.2725 / 0.2681  & 0.4364 / 0.4297 \\
      
    PPT~\cite{ref_22}& 
      0.1339 / 0.1309 & 11.3801 / 11.3499 & 68.78 / 50.50 & -8.8844 / -13.0090 & 56.95 / 61.96 & 0.2597 / 0.2992 & 0.5298 / 0.5469 & 
      0.3476 / 0.3268 & 16.5275 / 15.7775 & 20.36 / 30.08 & 0.8693 / 1.6747 & 70.91 / 196.25 & 0.3037 / 0.3399 & 0.5376 / 0.6001 \\

      % PPT$^\dag$& 
      % - & - & - & - & - & - & - & 
      % - & - & - & - & - & - & - \\

      {UMDST}~\cite{umdst} & 
      - / 0.1479 & - / 11.1785 & - / 65.40 & - / -4.1051 & - / 63.10 & - / 0.3065 & - / 0.6403
      & - / {\color{red}\textbf{0.5418}}  & - / 20.5546 & - / 54.60 & - / -1.5868 & - / 62.68 & - / 0.3360 & - / 0.5611\\

      % UMDST$^\dag$& 
      % - & - & - & - & - & - & - & 
      % - & - & - & - & - & - & - \\

      % GramGAN~\cite{gramgan}& 
      % - & - & - & - & - & - & - & 
      % - & - & - & - & - & - & - \\

      % GramGAN$^\dag$& 
      % - & - & - & - & - & - & - & 
      % - & - & - & - & - & - & - \\

    ASP~\cite{ref_7}& 
      0.1202 / 0.1219 & 10.9916 / 11.1032 & 61.00 / 66.95 & -9.9207 / -12.4642 & 66.75 / 69.59 & 0.2618 / 0.2582 & 0.5105 / 0.4978 & 
      0.4688 / 0.3929 & 19.5896 / 16.6227 & 75.66 / 76.03 & -0.2174 / 2.6554 & 39.36 / 66.36 & {\color{blue}\underline{0.2197}} / 0.2785 & {\color{red}\textbf{0.3751}} / 0.4975 \\

      % ASP$^\dag$& 
      % - & - & - & - & - & - & - & 
      % - & - & - & - & - & - & - \\

      TDKStain~\cite{ref_23}& 
      0.1275 / 0.1216 & 10.9003 / 11.1563 & 45.43 / 44.62 & -6.645 / 1.1292 & 66.39 / 92.79 & 0.2604 / 0.2512 & 0.5082 / 0.5040 & 
      0.4737 / 0.4788 & 19.5718 / 20.4984 & 43.83 / 43.63 & -0.8196 / -1.2804 & 56.46 / 59.66 & 0.2517 / 0.2559 & 0.4344 / 0.4304 \\

      % TDKStain$^\dag$& 
      % - & - & - & - & - & - & - & 
      % - & - & - & - & - & - & - \\

      SIMGAN~\cite{ref_25}& 
      0.1200 & 10.7020 & {\color{blue}\underline{76.90}} & {\color{red}\textbf{-0.1793}} & 58.32 & 0.2522 & 0.5196 & 
      0.4177 & 17.7036 & {\color{blue}\underline{80.79}} & 2.0767 & {\color{blue}\underline{37.93}} & 0.2357 & 0.4217 \\

      \cellcolor{gray!20}USI-GAN(ours)& 
      \cellcolor{gray!20}0.1267 & \cellcolor{gray!20}11.1342 & \cellcolor{gray!20}{\color{red}\textbf{81.98}} & \cellcolor{gray!20}-7.3421 & \cellcolor{gray!20}{\color{red}\textbf{48.54}} & \cellcolor{gray!20}{\color{red}\textbf{0.2345}} & \cellcolor{gray!20}{\color{red}\textbf{0.4611}} & 
      \cellcolor{gray!20}0.452 & \cellcolor{gray!20}19.4773 & \cellcolor{gray!20}{\color{red}\textbf{82.25}} & \cellcolor{gray!20}{\color{red}\textbf{0.0869}} & \cellcolor{gray!20}{\color{blue}\underline{37.01}} & \cellcolor{gray!20}{\color{red}\textbf{0.2139}} & \cellcolor{gray!20}{\color{blue}\underline{0.3761}} \\
      
      \bottomrule

       \hline

       \multirow{3}{*}{Method} & 
      \multicolumn{7}{c|}{Ki67} & \multicolumn{7}{c}{PR} \\
      \cmidrule(lr){2-15}
      & \multicolumn{2}{c|}{Quality (reference)} & \multicolumn{2}{c|}{Pathology} & \multicolumn{3}{c|}{Perception} & 
      \multicolumn{2}{c|}{Quality (reference)} & \multicolumn{2}{c|}{Pathology} & \multicolumn{3}{c}{Perception} \\
      \cmidrule(lr){2-15}
      & SSIM & PSNR & $R_{avg}\uparrow$ & IoD$^{\times10^7}$$\downarrow$ & FID$\downarrow$ & DISTS$\downarrow$ & PHV$_{avg}\downarrow$ & 
      SSIM & PSNR & $R_{avg}\uparrow$ & IoD$^{\times10^7}$ & FID$\downarrow$ & DISTS$\downarrow$ & PHV$_{avg}\downarrow$ \\
      \midrule
       \multicolumn{15}{c}{\textbf{GAN-based Image-to-Image Translation SOTA}} \\
      \midrule
      Pix2Pix~\cite{ref_14} & 
      0.4097 / 0.4309 & 20.3587 / {\color{red}\textbf{20.8746}} & 40.44 / 40.13 & -0.8362 / 1.0083 & 102.77 / 93.23 & 0.2837 / 0.2973 & 0.4887 / 0.5263 & 
      0.5038 / {\color{red}\textbf{0.6096}} & 23.083 / {\color{blue}\underline{24.9063}} & 26.61 / 26.05 & -0.475 / -0.8543 & 215.53 / 154.69 & 0.3072 / 0.3622 & 0.4837 / 0.5701 \\
      % Pix2Pix$^\dag$~\cite{ref_14} & 
      % - & - & - & - & - & - & - & 
      % - & - & - & - & - & - & - \\

      % PyramidP2P$^\dag$~\cite{ref_18} & 
      % - & - & - & - & - & - & - & 
      % - & - & - & - & - & - & - \\
      
      {Pix2PixHD}~\cite{wang2018pix2pixHD} & 
      {\color{red}\textbf{0.4279}} / 0.4155 & {\color{blue}\underline{20.3387}} / 19.8854 & 44.45 / 48.25 & 1.3103 / 1.8992 & 54.70 / 37.54 & 0.2586 / 0.2328 & 0.4830 / 0.4253 & 
      {\color{blue}\underline{0.5664}} / 0.5368 & 24.1064 / 23.4964 & 31.14 / 46.56 & -1.1365 / -0.9227 & 71.70 / 51.58 & 0.3114 / 0.2570 & 0.4821 / 0.4182 \\
      % Pix2PixHD$^\dag$ & 
      % - & - & - & - & - & - & - & 
      % - & - & - & - & - & - & - \\
      
      CycleGAN~\cite{ref_28} & 
      0.39 / 0.3911 & 19.2125 / 19.1552 & 81.94 / 82.35 & 0.0964 / 2.1075 & 39.47 / {\color{blue}\underline{25.74}} & {\color{blue}\underline{0.2073}} / 0.2444 & 0.3848 / {\color{red}\textbf{0.3807}} & 
      0.5546 / 0.565 & 23.2907 / 23.4863 & 54.22 / 49.74 & -0.5593 / -0.8913 & 43.98 / 37.25 & {\color{red}\textbf{0.2331}} / 0.2666 & 0.3647 / {\color{red}\textbf{0.3625}} \\

      % CycleGAN$^\dag$~\cite{ref_28} & 
      % - & - & - & - & - & - & - & 
      % - & - & - & - & - & - & - \\

     F-LseSim~\cite{ref_52} & 
      0.3856 / 0.3725 & 19.1239 / 19.1106 & 80.94 / 58.86 & 0.4046 / 2.377 & 37.69 / 32.89 & 0.2499 / 0.2538 & 0.4821 / 0.4762 & 
      0.4976 / 0.5269 & 20.4933 / 22.6819 & 47.18 / 43.12 & 0.4286 / -0.6384 & 68.77 / 50.43 & 0.3011 / 0.27788 & 0.4649 / 0.4450 \\

      % F-LseSim$^\dag$~\cite{ref_52} & 
      % - & - & - & - & - & - & - & 
      % - & - & - & - & - & - & - \\

      CUT~\cite{ref_19} & 
      0.3511 / 0.3738 & 19.0078 / 19.5456 & {\color{blue}\underline{82.11}} / 81.82 & 0.4491 / 2.2578 & 39.05 / 32.64 & 0.2424 / 0.2250 & 0.4841 / 0.4543 & 
      0.5452 / 0.5151 & 23.3178 / 22.3619 & 42.81 / 45.85 & -0.7435 / -0.2942 & 42.29 / 36.19 & 0.2646 / 0.2661 & 0.4095 / 0.4286 \\

      % CUT$^\dag$~\cite{ref_19} & 
      % - & - & - & - & - & - & - & 
      % - & - & - & - & - & - & - \\

      MoNCE~\cite{ref_26}& 
      0.3489 / 0.3571 & 18.5064 / 18.819 & 80.84 / 81.16 & 1.2873 / 2.4970 & 47.04 / 33.94 & 0.2495 / 0.2403 & 0.4955 / 0.4671 & 
      0.4845 / 0.4784 & 18.9298 / 21.3643 & 55.40 / 55.56 & 1.9725 / 0.6368 & 72.49 / 38.38 & 0.3174 / 0.2704 & 0.4805 / 0.4602 \\

      % MoNCE$^\dag$~\cite{ref_26}& 
      % - & - & - & - & - & - & - & 
      % - & - & - & - & - & - & - \\

      QSAttn~\cite{ref_40}& 
      0.384 / 0.355 & 19.8617 / 19.1989 & 81.69 / 81.86 & -0.2352 / 2.5348 & 36.89 / 44.19 & 0.2311 / 0.2353 & 0.4658 / 0.4711 & 
      0.5449 / 0.5168 & 23.3068 / 22.571  & 48.79 / {\color{blue}\underline{57.81}} & -0.3421 / {\color{red}\textbf{-0.0485}} & 37.01 / {\color{blue}\underline{35.89}} & 0.2540 / 0.2501 & 0.4119 / 0.4146 \\

      % QSAttn$^\dag$& 
      % - & - & - & - & - & - & - & 
      % - & - & - & - & - & - & - \\

      {NEGCUT}~\cite{wang2021instance}& 
      0.3528 / 0.3663 & 18.8025 / 19.3018 & 81.45 / 77.93 & 0.3715 / 2.1969 & 41.86 / 33.51 & 0.2410 / 0.2386 & 0.4829 / 0.4808 & 
      0.5095 / 0.4667 & 22.037 / 20.348 & 53.04 / 41.10 & 0.3156 / 0.7382 & 39.94 / 37.75 & 0.2612 / 0.2957 & 0.4466 / 0.47 \\

      % NEGCUT$^\dag$& 
      % - & - & - & - & - & - & - & 
      % - & - & - & - & - & - & - \\

    SRC~\cite{ref_50}& 
      - / 0.3489 & - / 18.6581 & - / 81.47 & - / 0.697 & - / 37.36 & - / 0.2369 & - / 0.4797 & 
      - / 0.4989 & - / 21.1762 & - / 54.68 & - / 0.8948 & - / 37.46 & - / 0.2750 & - / 0.4454 \\

      % SRC$^\dag$& 
      % - & - & - & - & - & - & - & 
      % - & - & - & - & - & - & - \\

      {DECENT}~\cite{xie2022unsupervised}& 
      0.3718 / 0.3602  & 19.2342 / 19.2451 & 80.75 / 78.65 & 0.1254 / 2.1868 & 30.70 / 33.08 & 0.2422 / 0.2442 & 0.4860 / 0.4942 & 
      0.5466 / 0.5104 & 23.2836 / 22.0259 & 55.48 / 54.28 & -0.7105 / -0.4417 & 40.26 / 40.42 & 0.2450 / 0.2659 & {\color{blue}\underline{0.3791}} / 0.4529 \\

      % DECENT$^\dag$& 
      % - & - & - & - & - & - & - & 
      % - & - & - & - & - & - & - \\

      % SANTA~\cite{xie2023unpaired}& 
      % 0.3407 & 19.5842 & 79.98 & 0.4762 & {\color{blue}\underline{23.82}} & 0.2097  & 0.42265 & 
      % 0.5031 & 23.8335 & 53.98 & -16.069 & 36.27 & 0.2193 & 0.4012 \\

      % SANTA$^\dag$& 
      % - & - & - & - & - & - & - & 
      % - & - & - & - & - & - & - \\

     {EnCo}~\cite{cai2024rethinking}& 
      0.3917 / 0.3751 & 19.4617 / 19.158 & 78.53 / 79.05 & 2.1865 / 2.4694 & 38.648 / 38.17 & 0.2341 / 0.2377 & 0.46315 / 0.4704 & 
      0.5038 / 0.5113 & 20.8565 / 22.5336 & 19.83 / 43.42 & -0.3983 / -0.4417 & 66.36 / 53.41 & 0.3330 / 0.2643 & 0.5234 / 0.4420 \\

      % EnCo$^\dag$& 
      % - & - & - & - & - & - & - & 
      % - & - & - & - & - & - & - \\

    % StegoGAN~\cite{wu2024stegogan}& 
    %   - & - & - & - & - & - & - & 
    %   - & - & - & - & - & - & - \\

      % StegoGAN$^\dag$& 
      % - & - & - & - & - & - & - & 
      % - & - & - & - & - & - & - \\
    \midrule
       \multicolumn{15}{c}{\textbf{Diffusion-based Image-to-Image Translation SOTA}} \\
      \midrule
    
      {UNSB}~\cite{ref_51}& 
      0.3481 / - & 20.5063 / - & 78.34 / - & -1.9469 / - & 31.81 / - & 0.2109 / - & 0.4611 / - & 
      0.511 / - & 23.7755 / - & 56.31 / - & -1.7314 / - & 37.01 / - & {\color{blue}\underline{0.2442}} / - & 0.4502 / - \\

   {DDBM}~\cite{ddbm}& 
      0.3567 / - & 18.8877 / - & 44.71 / - & 2.7059 / - & 109.45 / - & 0.3013 / - & 0.5659 / - & 
      0.5043 / - & 22.293 / - & 11.78 / - & -1.0088 / - & 234.86 / - & 0.3171 / - & 0.5499 / - \\

      {ControlNet}~\cite{ref_controlnet}& 
      0.2035 / - & 8.7388 / - & 37.32 / - & 10.4722 / - & 246.05 / - & 0.4117 / - & 0.7179 / - & 
      0.2442 / - & 8.2439 / - & 0.093 / - & 21.7282 / - & 260.62 / - & 0.4449 / - & 0.7148 / - \\
      {VIMs}\cite{dubey2024vimsvirtualimmunohistochemistrymultiplex}& 
      0.4015 / - & 20.8459 / - & 55.17 / - & 0.5972 / - & 80.61 / - & 0.2986 / - & 0.5371 / - & 
      0.573 / - & 24.4844 / - & 32.63 / - & -1.5848 / - & 109.73 / - & 0.3090 / - & 0.4629 / - \\

    \midrule
       \multicolumn{15}{c}{\textbf{Virtual Staining SOTA}} \\
      \midrule

    PyramidP2P~\cite{ref_18} & 
      0.39 / 0.4042 & 20.2169 / 20.3703 & 48.48 / 47.02 & -0.7704 / 1.3461 & 57.24 / 55.18 & 0.2365 / 0.2581  & 0.4263 / 0.4666 & 
      0.5258 / 0.5225 & 23.9905 / 24.2178 & 41.01 / 35.67 & -1.1295 / -1.1742 & 62.83 / 95.60 & 0.2576 / 0.2782 & 0.3951 / 0.4161 \\
      
    PPT~\cite{ref_22}& 
      0.3911 / 0.2825 & 19.5607 / 16.7756 & 72.49 / 62.48 & 0.91 / 2.7694 & 54.91 / 102.98 & 0.2211 / 0.2556 & 0.4539 / 0.5454 & 
      0.529 / 0.2968 & 23.3723 / 18.4435 & 36.83 / 16.55 & -1.1287 / 0.2280 & 68.88 / 322.23 & 0.2676 / 0.3442 & 0.4240 / 0.6314 \\

      % PPT$^\dag$& 
      % - & - & - & - & - & - & - & 
      % - & - & - & - & - & - & - \\

      {UMDST}~\cite{umdst}& 
      - / 0.4734  & - / 20.1886 & - / 43.21 & - / 1.0701 & - / 56.41 & - / 0.2898 & - / 0.5723 & 
      - / 0.5863 & - / 22.5572 & - / 48.92 & - / {\color{blue}\underline{-0.0899}} & - / 44.76 & - / 0.3014 & - / 0.5216 \\

      % UMDST$^\dag$& 
      % - & - & - & - & - & - & - & 
      % - & - & - & - & - & - & - \\

      % GramGAN~\cite{gramgan}& 
      % - & - & - & - & - & - & - & 
      % - & - & - & - & - & - & - \\

      % GramGAN$^\dag$& 
      % - & - & - & - & - & - & - & 
      % - & - & - & - & - & - & - \\

    ASP~\cite{ref_7}& 
      0.3743 / 0.3893 & 19.1065 / 19.3626 & 75.22 / 74.95 & 0.2612 / 1.8991 & 35.86 / 34.31 & 0.2334 / 0.2233 & 0.4475 / 0.4347 & 
      0.5762 / 0.5654 & {\color{blue}\underline{24.4079}} / 24.2786 & 51.61 / 53.16 & -1.2598 / -1.2196 & 67.65 / 63.55 & 0.2522 / 0.2449 & 0.4137 / 0.3957 \\

      % ASP$^\dag$& 
      % - & - & - & - & - & - & - & 
      % - & - & - & - & - & - & - \\

      TDKStain~\cite{ref_23}& 
      {\color{blue}\underline{0.4138}} / 0.4006 & 20.0043 / 19.4149 & 45.86 / 46.03 & -0.6335 / -8.5305 & 57.74 / 43.57 & 0.2411 / 0.2225 & 0.4185 / 0.4194 & 
      0.4998 / 0.5416 & 23.3719 / 23.3434 & 15.82 / 16.48 & -1.1883 / -0.9537 & 107.13 /63.78 & 0.2590 / 0.2590  & 0.4329 / 0.4135 \\

      % TDKStain$^\dag$& 
      % - & - & - & - & - & - & - & 
      % - & - & - & - & - & - & - \\

      SIMGAN~\cite{ref_25}& 
      0.3408 & 18.8144 & 81.80 & {\color{blue}\underline{0.1395}} & 38.37 & 0.2397 & 0.4934 & 
      0.5247 & 21.999 & 45.23 & -0.3719 & 40.05 & 0.2718 & 0.4396 \\

      \cellcolor{gray!20}USI-GAN(ours)& 
      \cellcolor{gray!20}0.3773 & \cellcolor{gray!20}19.209 & \cellcolor{gray!20}{\color{red}\textbf{82.23}} & \cellcolor{gray!20}{\color{red}\textbf{0.009}} & \cellcolor{gray!20}{\color{red}\textbf{23.74}} & \cellcolor{gray!20}{\color{red}\textbf{0.2071}} & \cellcolor{gray!20}{\color{blue}\underline{0.4064}} & 
      \cellcolor{gray!20}0.2163 & \cellcolor{gray!20}14.2176 & \cellcolor{gray!20}{\color{red}\textbf{61.80}} & \cellcolor{gray!20}3.2447 & \cellcolor{gray!20}{\color{red}\textbf{34.63}} & \cellcolor{gray!20}0.3499 & \cellcolor{gray!20}0.4559 \\
      \bottomrule
    \end{tabular}
  }
\end{table*}

\begin{figure*}[!h]
\centerline{\includegraphics[width=\textwidth]{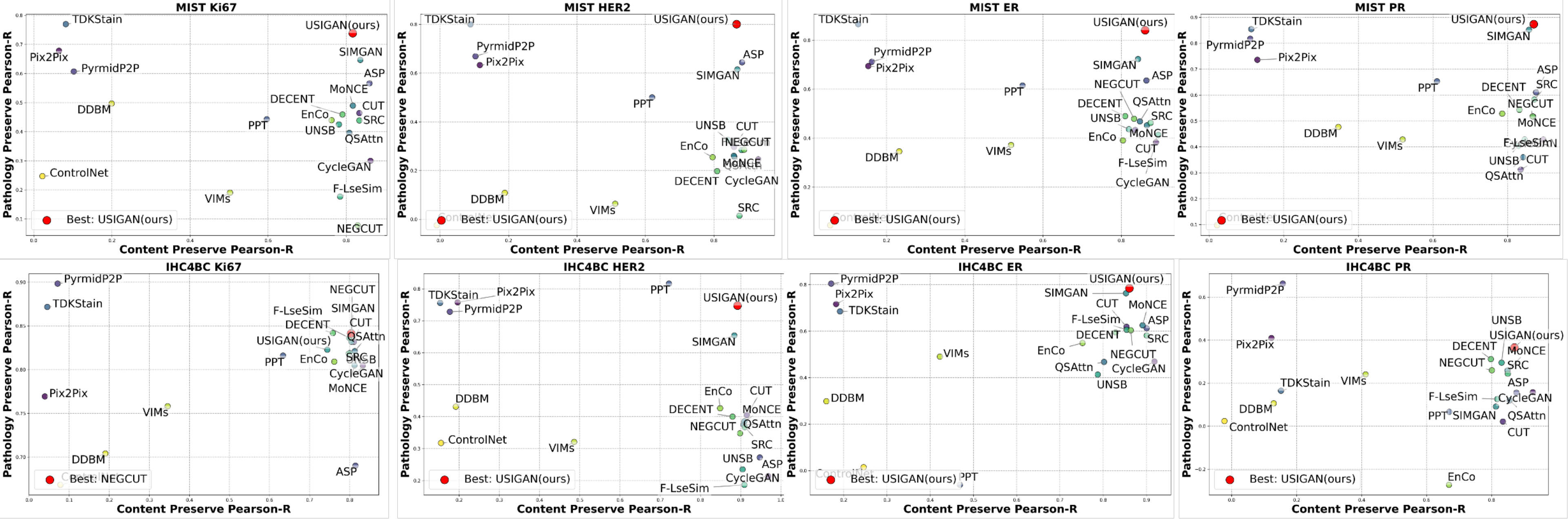}}
\caption{The visualization results of pathological semantic consistency and content preservation across different methods demonstrate that our approach exhibits outstanding performance in both aspects.}
\label{fig_perasonr}
\end{figure*}

\begin{table*}[ht]
    \centering
    \caption{Comparison of stain transfer performance for different methods on the MIST benchmark datasets. The best and second-best scores are in \textbf{bold} and \underline{underline}, respectively. IoD,FID and PHV$_{avg}$ (The lower is better.)}
    \label{tab:all_stain_comparison}
    \resizebox{\textwidth}{!}{
    \begin{tabular}{lcccccc|cccccc}
        \toprule
        \multirow{2}{*}{Methods} & \multicolumn{6}{c}{Ki67} & \multicolumn{6}{c}{ER} \\
        \cmidrule(lr){2-7} \cmidrule(lr){8-13}
        & {IoD}$\downarrow$ & {$R_{c}(\%)$}$\uparrow$  & {$R_{p}(\%)$}$\uparrow$ & {$R_{avg}(\%)$}$\uparrow$ & {FID}$\downarrow$ & {PHV$_{avg}$}$\downarrow$ 
        & {IoD}$\downarrow$ & {$R_{c}(\%)$}$\uparrow$ & {$R_{p}(\%)$}$\uparrow$ & {$R_{avg}(\%)$}$\uparrow$ & {FID}$\downarrow$ & {PHV$_{avg}$}$\downarrow$ \\
        \midrule
        Baseline         & 0.4895 & 77.86 & 36.34    & 57.10      & \underline{27.45}      & 0.5023  & 0.8598 & 80.71  & 42.18     & 61.44      & 35.27      & 0.5103 \\
        w/o UOT-CTM      & \underline{0.2595}  & 80.75 & \underline{71.81}     & \underline{76.28}      & 29.40     & 0.5024 & 1.8404     & 83.65      & \underline{81.91}      & \textbf{82.78}  & 35.35      & 0.4662 \\
        w/o PC-SCM       & \textbf{0.1915}  & \textbf{82.70} & 44.68     & 63.69      & 30.59     & \underline{0.4977}      & \textbf{0.4649}      & \textbf{86.35}   & 46.31   & 66.33 & \textbf{32.84}      & 0.4991 \\
        USI-GAN (ours)   & 0.2680  & \underline{81.63} & \textbf{73.88}     & \textbf{77.75}      & \textbf{27.36}     & \textbf{0.4702} & \underline{0.5109}     & \underline{86.06}      & \textbf{84.07} & \textbf{85.07}     & \underline{33.06} & \underline{0.4694} \\
        \midrule
        \multicolumn{9}{c}{} \\ % 添加空行分隔
        \toprule
        \multirow{2}{*}{Methods} & \multicolumn{6}{c}{PR} & \multicolumn{6}{c}{HER2} \\
        \cmidrule(lr){2-7} \cmidrule(lr){8-13}
        & {IoD}$\downarrow$ & {$R_{c}(\%)$}$\uparrow$ & {$R_{p}(\%)$}$\uparrow$ & {$R_{avg}(\%)$}$\uparrow$ & {FID}$\downarrow$ & {PHV$_{avg}$}$\downarrow$ 
        & {IoD}$\downarrow$ & {$R_{c}(\%)$}$\uparrow$ & {$R_{p}(\%)$}$\uparrow$ & {$R_{avg}(\%)$}$\uparrow$ & {FID}$\downarrow$ & {PHV$_{avg}$}$\downarrow$ \\
        \midrule
        Baseline         & 2.4654  & 82.06 & 25.24     & 53.65      & 37.97     & 0.5220 & 1.7549     & 85.32      & 32.80      & 59.06 & 42.14 & 0.5082 \\
        w/o UOT-CTM      & 1.6829  & 83.53 & \underline{86.73}     & \underline{85.13}      & 43.40     & 0.4932 & \textbf{0.7819}     & 84.32      & \textbf{82.94}      & \textbf{83.63} & \textbf{37.39}      & 0.4728  \\
        w/o PC-SCM      & \textbf{0.4736}  & 87.19 & 48.90     & 68.05      & 34.26     & 0.5150 & 1.5598     & \underline{85.81}      & 21.39      & 53.60 & 42.09      & 0.5247 \\
        USI-GAN (ours)   & 2.0700  & \textbf{87.03} & \textbf{87.37}     & \textbf{87.20}      & \underline{34.64}     & \textbf{0.456} & 1.6788     & \textbf{86.63}      & \underline{80.08}     & \underline{83.35} & \underline{37.76} & \textbf{0.4708} \\
        \bottomrule
    \end{tabular}}
\end{table*}

\subsubsection{Results on MIST}

{{Table \ref{tab:mist_comparsion}} presents the quantitative experimental results on the MIST dataset. Our USIGAN demonstrates significant advantages in both clinical value and image quality. We evaluated our method using three categories of metrics: pixel-level, feature-level, and fluorescence intensity. Considering the instability of GAN models and the observation that our method performs better with batch sizes greater than 1 compared to other methods, we also compared the results for a batch size of 1 and results aligned with the batch size used in our method. The results indicate that, compared to supervised learning methods, a batch size of 1 provides noticeable advantages. However, due to the influence of weak pairing, the generated results exhibit lower content consistency. The IoD metric evaluates the difference in total fluorescence intensity between the test set and the adjacent real slices. Ideally, combining IoD with $R_{avg}$ can provide insights into the similarity of virtual IHC staining to its real pathological diagnostic significance. As shown in Figure \ref{fig_e1_1}, although supervised learning methods like Pix2Pix achieve high pathological semantic relevance, virtual IHC generated by these methods often suffers from noticeable artifacts and distortions, which limit their practical applicability. Therefore, we believe that $R_{avg}$ provides a more comprehensive evaluation of their clinical value. On the MIST dataset, our method outperforms existing approaches in perceptual and pathological semantic metrics, demonstrating superior performance and clinical relevance.}

\subsubsection{Results on IHC4BC}

{{Table \ref{tab:ihc4bc_comparsion}} presents the quantitative experimental results on the IHC4BC dataset. It is important to note that in the PR subset of the IHC4BC dataset, the pathological images exhibit low contrast in non-positive regions, appearing almost transparent. This makes it easier to distinguish positive regions based on optical density (OD) values. Due to this characteristic, the significant contrast in positive regions of the PR subset makes global features, such as boundary definition and contrast intensity, more critical during evaluation.}

{Our method, which binds optical density (OD) to pathological semantic information, is highly sensitive to positive regions and tends to classify cells as positive. While this is advantageous in datasets with abundant positive regions, it poses a challenge in datasets with fewer abnormal regions. As shown in Figure \ref{fig_perasonr}, apart from supervised methods, the pathological semantic relevance of all methods is significantly lower on this dataset compared to others. Nevertheless, our method continues to demonstrate outstanding performance. }

\begin{table}[ht]
    \centering
    \caption{The ablation experiments on MIST ER for all modules are conducted, where \(\mathcal{L}_{\text{tcyc}}\) represents the UOT-CTM module, and the remaining columns represent the PC-SCM module.}
    \label{tab:all_module}
    \resizebox{\columnwidth}{!}{
    \begin{tabular}{ccccccccc}
        \toprule
         $\mathcal{L}_{odc}$  & $\mathcal{L}_{cc}$ & $\mathcal{L}_{tcyc}$  & $R_c(\%)\uparrow$ & $R_p(\%)\uparrow$ &  $R_{avg}(\%)\uparrow$  & FID $\downarrow$ & {PHV$_{avg} \downarrow$} \\
        \midrule
         --  & -- & -- & 80.71 & 42.18 & 61.44 & 35.27 & 0.5103 \\ % baseline
        % -- & \checkmark & -- & -- & -- & -- & -- & -- & -- & -- & - \\ % scm
         % --  & \checkmark & -- & -- & -- & -- & -- & -- & -- & --\\ % scm+od
          \checkmark  & -- & -- & 84.06 & 81.54 & \underline{82.80} & 35.16 & 0.4776\\ % scm+od+avg
          --  & \checkmark & -- & 82.63 & 75.55 & 79.09 & \textbf{32.67} & 0.4908\\ % baseline+cc
          \checkmark  & \checkmark & - & 83.65 & 81.91 & 81.01 & 35.35 & \textbf{0.4662} \\ % cc+odc
          --  & -- & \checkmark & \textbf{86.35} & {46.30} & 66.33 & \underline{32.84} & {0.4991}\\ % baseline+ uot
         % UOT+OT  & --  & -- & \checkmark & {85.91} & {46.08} & {0.2952} & {32.52} & {0.5068}\\ % uot_cyV2
% \rowcolor{red!20}          OT(SIM-GAN)\cite{ref_25}  & --  & \checkmark & \times ($\mathcal{L}_{1}$) & 84.26 & 72.28 & 1.4809 & 34.61 & 0.4977 \\ % SIM-GAN
           --  & \checkmark & \checkmark & 85.88 & 70.85 & 78.36 & 33.05 & 0.4846\\ % uot_cy_cc
        % UOT+OT  & --  & \checkmark & \checkmark & {87.34} & {72.55} & {1.3640} & {34.41} & {0.4841} \\ % UOT_cyV2_CC
          \checkmark & --  & \checkmark & 85.62 & 76.58 & 81.10 & 33.25 & 0.5049  \\
         % UOT+OT  & \checkmark & -- & \checkmark & -- & -- & -- & -- & -- \\
         
         \midrule
\rowcolor{gray!20}          \checkmark  & \checkmark & \checkmark & \underline{86.06} & \textbf{84.07} & \textbf{85.07} & 33.06 & \underline{0.4694}   \\ %uot_cy_cc_pcscm
         
% \rowcolor{gray!20}         UOT+OT  & \checkmark & \checkmark & \checkmark & -- & -- & -- & -- & --  \\ %UOT_cyV2_PCSCM
        \bottomrule
    \end{tabular}}
\end{table}

\begin{table}[!htb]
    \centering
    \caption{The quantitative comparison results of the effects of UOT and UOT-CTM on USIGAN.  $\mathcal{L}_1$ w (U)OT represent constrain transport strategy between $T^{(H-I)}$ and $T^{(I-I)}$, and (U)OT represents the theory for computing the cost matrix of feature matching.}
    \label{tababalation}
    \resizebox{\columnwidth}{!}{
    \begin{tabular}{cccccc} % 所有列居中对齐
        \toprule
        {Strategy} & {R$_c(\%) \uparrow$} & {R$_p(\%)\uparrow$} & {IoD$\downarrow$} & {FID$\downarrow$} & {PHV$_{avg} \downarrow$}  \\
        \midrule
        baseline & 80.71 & 42.18 & 0.8590 & 35.27 & 0.5103 \\
        $\mathcal{L}_{1}$ w OT & 86.19 & 37.51 & 2.5027 & 33.64 & 0.5243\\
        $\mathcal{L}_{1}$ w UOT & 85.69 & 43.61 & 2.2763 & 33.03 & 0.5042\\
        % OT-SIM\cite{ref_25}+ $\mathcal{L}_{cyc}$ & 86.70 & 42.83 & 2.5027 & 35.56  & 0.5301\\
        \midrule
\rowcolor{gray!20}        $\mathcal{L}_{tcyc}$(OT) & 86.25 & 40.51 & 1.4186 & 33.23 & 0.5084\\
        % $\mathcal{L}_{tcyc}$(UOT-CTM) & \textbf{86.35} & \textbf{46.30} & \textbf{0.4640} & \textbf{32.84} & \textbf{0.4991}\\
\rowcolor{gray!20}$\mathcal{L}_{tcyc}$(UOT) & \textbf{86.35} & \textbf{46.30} & \textbf{0.4640} & \textbf{32.84} & \textbf{0.4991}\\
% \rowcolor{gray!20}$\mathcal{L}_{tcyc}$(UOT+OT) & {85.91} & {46.08} & \textbf{0.2952} & \textbf{32.52} & {0.5068}\\
        % OT-SIM+$\mathcal{L}_{cc}$\cite{ref_25} & 84.26 & 72.28 & 1.4809 & 34.61 & 0.4977\\
        % OT-CTM+$\mathcal{L}_{cc}$ & 85.79 & 73.31 & 0.5326 & 33.62 & 0.4933\\
        % \midrule
        % UOT-SIM+ $\mathcal{L}_{tcyc}$ & 87.37 & 46.91 & 0.8664 & 34.57 & 0.5217\\
        
        % UOT-CTM+$\mathcal{L}_{cc}$ & 85.88 & 70.85 & \textbf{0.4216} & 33.05 & \textbf{0.4846}\\
       
        % w/o SCM & - & - & - & - & -\\
       
        \bottomrule
    \end{tabular}}
\end{table}

% \begin{figure}[htb]
% \centerline{\includegraphics[width=\columnwidth]{uot-simgan.pdf}}
% \caption{Qualitative and quantitative analyses are conducted by directly replacing classical optimal transport theory with unbalanced optimal transport on SIMGAN \cite{ref_25}.}
% \label{fig_simgan_uot}
% \end{figure}

\begin{figure}[htb]
    \centering
    \includegraphics[width=\linewidth]{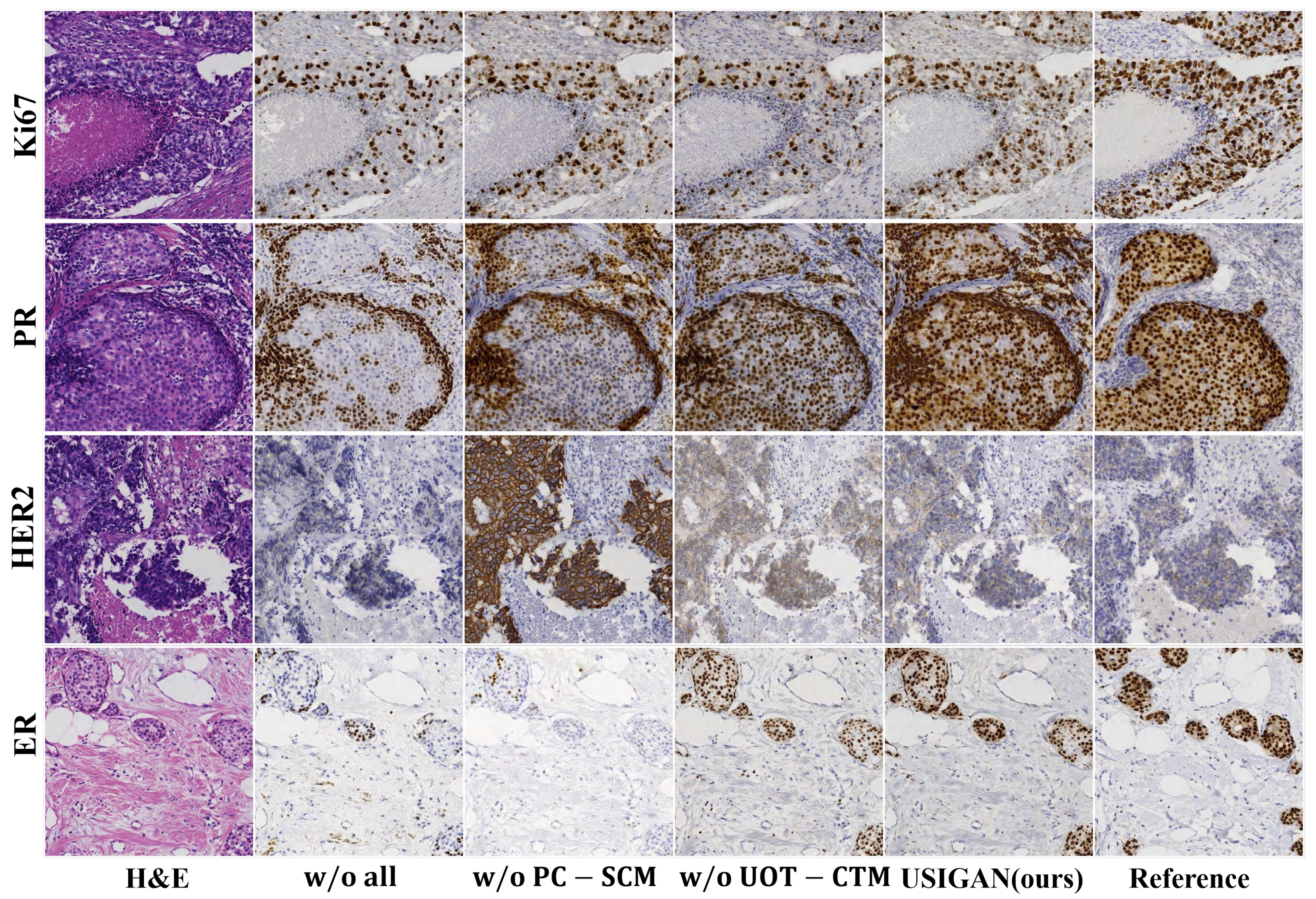}
    \caption{Visualize the impact of UOT-CTM and PC-SCM on USIGAN on MIST dataset benckmark.}
    \label{fig_all_abalation}
\end{figure}

\begin{figure}[htb]
\centerline{\includegraphics[width=\columnwidth]{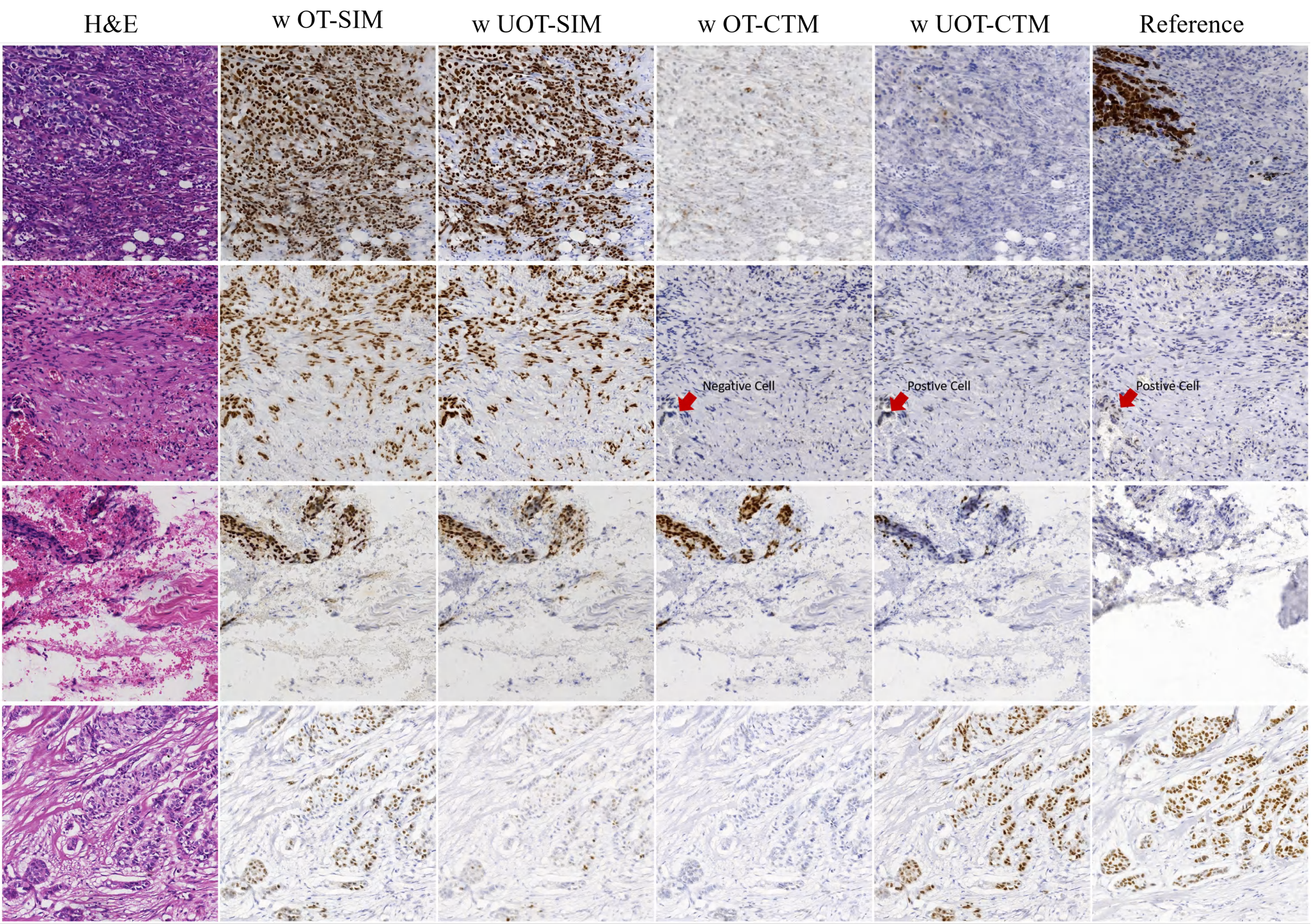}}
\caption{{Visualize the impact of UOT and transport consistency loss compared to the $\mathcal{L}_1$ loss that directly constrains both transport matrix.}}
\label{fig_uot_ctm}
\end{figure}

\begin{table}[!ht]
    \centering
    \caption{The quantitative results of the influence of different batch size on the performance of the PC-SCM Mechanism.}
    % \resizebox{\columnwidth}{!}{
    \begin{tabular}{ccccccc}
        \toprule
         Batch size   & $R_c(\%)$ & $R_p(\%)$ & IoD & FID & {PHV$_{avg}$}  \\
        \midrule
         1   &84.58 & 23.61 & 1.9810  & 47.95 & 0.5337\\ % 
        2   &84.19 & 57.95 & 0.8635 & 49.56 & 0.5136\\ % 
        4   & 85.23 & 57.11 & 2.1414 & 49.18 & 0.5302\\ % 
        8   & {86.14} & 57.90 & 2.6316  & 57.81 & 0.5245\\ % 
        16   & 82.79 & 61.70 & 4.0556 & 64.54 & 0.5480 \\ % 
        \bottomrule
    \end{tabular}%}
    \label{tab:bsab}
\end{table}

% \begin{figure*}[!ht]
%     \centering
%     \includegraphics[width=\linewidth]{loss_curve.png}
%     \caption{The training loss graphs of USI-GAN for each sub-loss of Ki67, ER, PR, and HER2 virtual staining on the MIST dataset.}
%     \label{fig:loss_curve}
% \end{figure*}

\begin{figure*}[!ht]
\centerline{\includegraphics[width=0.9\textwidth]{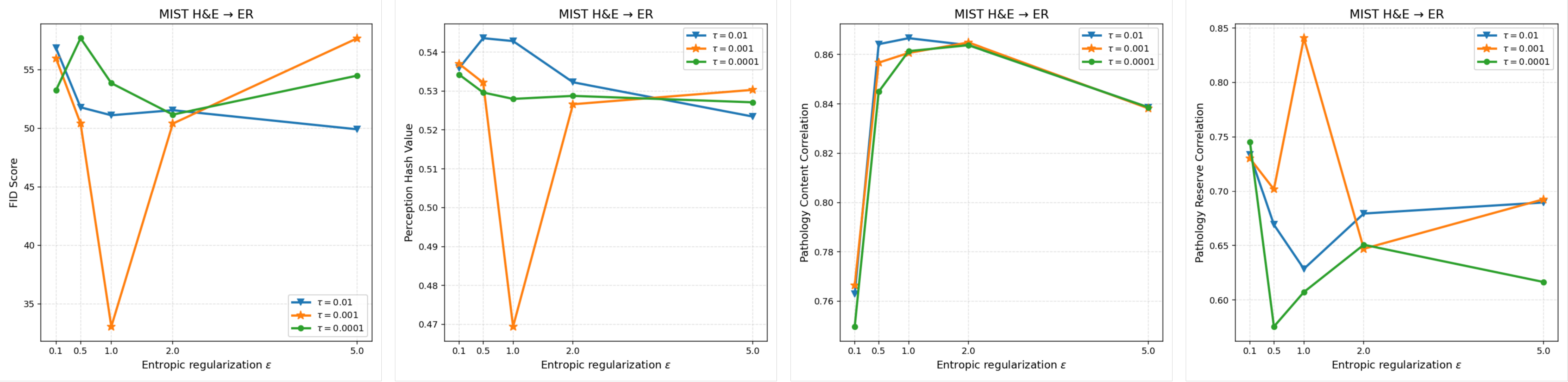}}
\caption{{Ablation study  for the unbalanced optimal transport hyperparameters: entropic regularization $\epsilon$ and relaxation coefficient $\tau$ in the UOT-CTM module when calculating transport costs for MIST ER staining.}}
\label{fig_taueps_1}
\end{figure*}

% \begin{figure}[!ht]
%     \centering
%     \includegraphics[width=\linewidth]{pearson_batch_size.png}
%     \caption{The infulenced different batch size  on the Pearson Correlation coefficient between virtual IHC images and adjacent layer real IHC images on PC-SCM mechanism}
%     \label{fig:perason}
% \end{figure}

\begin{table}[!h]
    \centering
    \caption{The quantitative results of the different relax parameter $\tau$ on the performance of the UOT-CTM Mechanism on MIST ER.}
    % \resizebox{\columnwidth}{!}{
    \begin{tabular}{ccccccc}
        \toprule
        $\tau$   & $R_c(\%)$ & $R_p(\%)$ & IoD & FID & {PHV$_{avg}$}  \\
        \midrule
         0.0001   & 86.63 & 52.81 & 0.4639  & 57.39 & 0.5642\\ % 
        0.001   & 86.35 & 46.31 & 0.4649 & 32.84 & 0.4991\\ % 
        0.01   & 87.00 & 48.41 & 0.5813 & 59.18 & 0.5772\\ % 
        0.1   & 86.12 & 42.69 & 0.5827  & 58.66 & 0.5735\\ % 
        \bottomrule
    \end{tabular}%}
    \label{fig:tauab}
\end{table}

\begin{figure}[!ht]
    \centering
    \includegraphics[width=\linewidth]{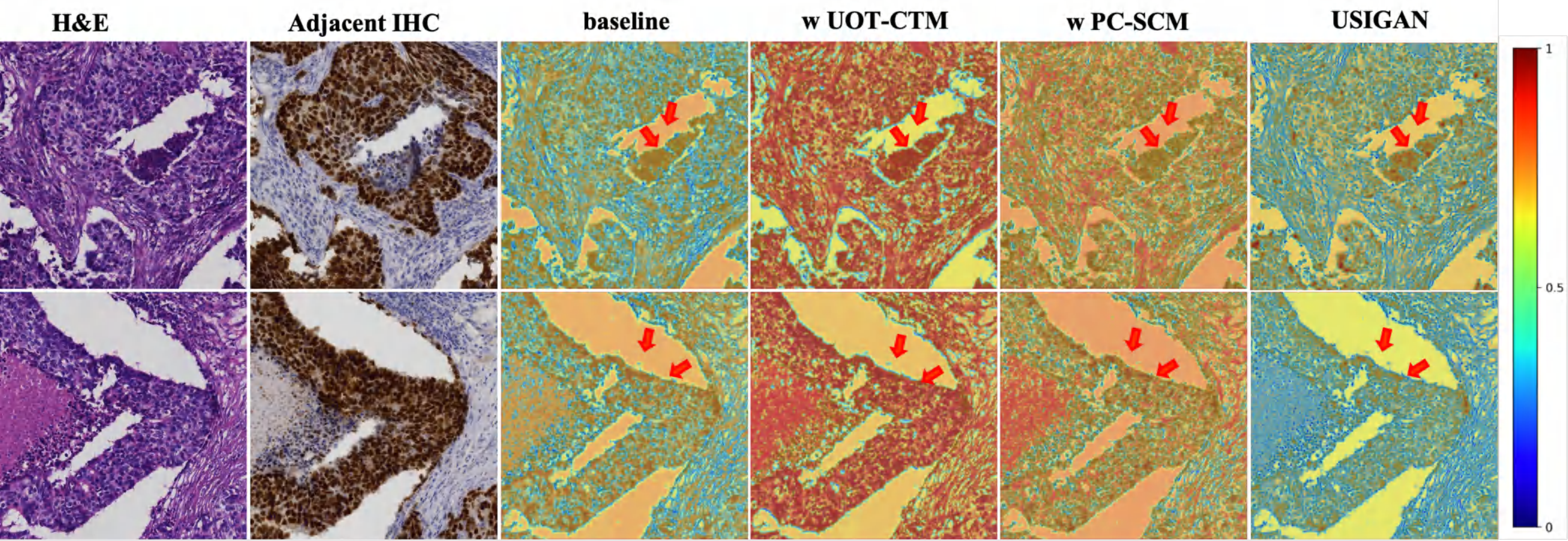}
    \caption{{The ablation study visualizations of model attention region changes demonstrate that models with self-information mining gradually enhance the contrast between attention to key foreground regions and the background.}}
    \label{fig:selfinfo}
\end{figure}

\subsection{Ablation Study and Analysis}

{To investigate the effectiveness of the cyclic transfer strategies UOT-CTM and PC-SCM, we measured content consistency using the Pearson-R coefficient $R_c$, pathological semantic consistency using $R_p$, and overall clinical value using $R_{{avg}}$. As shown in {Table~\ref{tab:all_module}} and {Table~\ref{tababalation}}, incorporating the UOT-CTM strategy represented by $\mathcal{L}_{tcyc}$ significantly improved content consistency, while the improvement in pathological semantic consistency was less pronounced. On the other hand, the PC-SCM mechanism substantially enhanced pathological semantic consistency but performed poorly in maintaining content consistency. By combining both modules, our method leverages the advantages of each, achieving a balance between content and pathological semantic consistency compared to the baseline.}

{Additionally, to explore the effect of optical density anchoring on pathological semantics, we analyzed rows 2, 3, and 4 in {Table~\ref{tab:all_module}}. The multi-level feature consistency loss $\mathcal{L}_{cc}$ improves pathological representation consistency by leveraging features across multiple levels. However, due to the inconsistency of low-level features such as morphology and texture in weakly paired data, $\mathcal{L}_{cc}$ reintroduces the influence of weak pairing errors. In contrast, the optical density consistency loss $\mathcal{L}_{odc}$ directly anchors pathological semantics, showing a more significant improvement in pathological semantic consistency $R_p$ compared to $\mathcal{L}_{cc}$, but at the cost of slightly lower image quality as measured by FID. Therefore, we adopted $\mathcal{L}_{cc}$ as an auxiliary term, combining it with $\mathcal{L}_{odc}$ in the PC-SCM strategy. This combination led to significant improvements in perceptual metrics, as illustrated by the visual results in {Figure~\ref{fig_all_abalation}}.}

{To further discuss the effectiveness of eliminating the impact of weakly paired data and the limitations of classical optimal transport (OT) in such scenarios, we analyzed the differences between directly constraining the transport matrices $T^{(H-I)}$ and $T^{(H-\hat{I})}$ using $\mathcal{L}_1$, without eliminating weakly paired terms, and employing the cyclic transfer strategy $\mathcal{L}_{tcyc}$. Additionally, we compared the differences in transport cost computation between unbalanced optimal transport (UOT) and classical OT. As shown in {Figure~\ref{fig_uot_ctm}} and {Table~\ref{tababalation}}, applying OT to compute transport costs without eliminating weakly paired terms resulted in a decrease in pathological semantic consistency $R_p$. Under cyclic transfer, using OT for transport cost computation also led to a lower $R_p$ compared to the baseline, though it performed better than the strategy without eliminating weakly paired terms. When using UOT to compute transport costs, the cyclic transfer strategy achieved optimal results across all metrics. This demonstrates that the cyclic transfer loss $\mathcal{L}_{tcyc}$ can better learn both content and style information in the transport matrices, avoiding a trade-off between the two. Furthermore, UOT provides a more accurate transport cost matrix, enhancing the overall performance.}

\subsubsection{Sensitivity to batch size on PC-SCM}

PC-SCM utilizes intra-batch correlation to guide the generated results in maintaining pathological consistency across batches. Therefore, the batch size influences the effectiveness of PC-SCM. To explore the impact of batch size, we set the image reading size to 256 and randomly crop it to 128 for easier requirement of memory. As shown in Table \ref{tab:bsab} The Optical Density  vectors, compressed through Focal Optical Density , exhibit increasing instability as the batch size grows. This phenomenon arises due to the weaker direct correlation between morphological features and the intensity values of optical density. As a result, relatively satisfactory results can be achieved at smaller batch sizes, where the impact of this instability is minimized. Overall, a batch size of 4 achieves the best performance by balancing the trade-off between morphological consistency and self-information mining. However, due to the higher computational cost associated with larger batch sizes, we adopt a batch size of 2 as our training configuration. This choice effectively reduces training overhead while maintaining a satisfactory level of performance.

\subsubsection{Sensitivity to $\tau$/$\epsilon$ on UOT-CTM}

{The UOT-CTM relies on Sinkhorn-Knopp entropy regularization and KL divergence to achieve optimal transport and relaxed transport boundaries. The relaxation coefficient $\tau$ and the entropy regularization parameter $\epsilon$ directly affect the quality of the transport cost matrix.
The parameter $\epsilon$ adds a regularization term that impacts the smoothness of the dual variables in optimal transport. When $\epsilon$ is small, the transport plan is closer to the original cost matrix and is more sparse; however, when $\epsilon$ is large, the transport plan becomes smoother but may risk losing fine details. The parameter $\tau$ determines whether the target mass in the transport plan strictly matches the source mass. In weakly paired datasets with varying matching rates, the optimal $\tau$ can vary. When $\tau$ is too small, unbalanced optimal transport approaches traditional optimal transport, while excessively large $\tau$ may lead to numerical instability.}
{Table \ref{fig:tauab} and Figure \ref{fig_taueps_1} illustrate the results of virtual staining with different $\tau$ and $\tau/\epsilon$ combinations for the ER staining type on the MIST dataset. When $\epsilon$ is too large and $\tau$ is too small, the transport plan becomes overly sparse and numerically unstable. Conversely, when $\epsilon$ is too small and $\tau$ is too large, the transport plan is smooth but loses too much detail, resulting in poorer content consistency but good pathological semantic consistency.
For all datasets, we selected $\tau=0.001$ and $\epsilon=1$ as the default settings. However, we believe that more optimal $\tau/\epsilon$ combinations may exist for datasets with different matching rates.}

% \subsubsection{Parameter ablation}

\begin{figure}
    \centering
    \includegraphics[width=\linewidth]{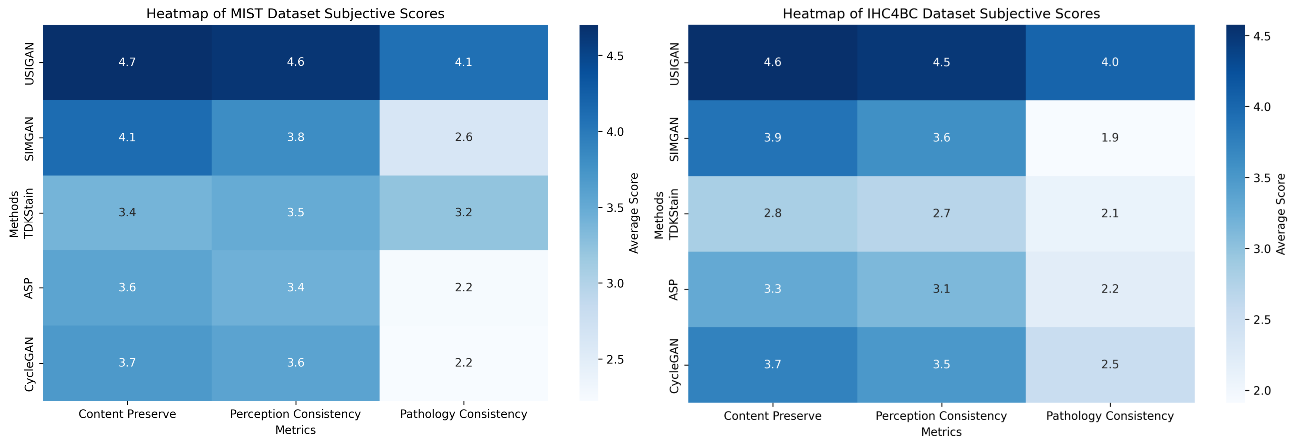}
    \caption{{The heatmap below represents the subjective evaluation scores of pathologists for Immunohistochemistry (IHC) images. These scores were assigned based on their assessment of diagnostic clinical value.}}
    \label{fig:subjectiv}
\end{figure}

\subsection{Subjetive Evaluation}

{We invited three pathologists to conduct a double-blind subjective evaluation. For each staining type, we randomly selected 10 images for each method, resulting in a total of 400 images from the MIST and IHC4BC datasets. These images were presented in randomized order, showing the H\&E image alongside the corresponding adjacent IHC and virtual IHC for scoring. Scores were assigned on a scale of 0 to 5 across three dimensions: (1) structural fidelity, (2) visual quality, and (3) diagnostic usability. }

{As shown in Figure~\ref{fig:subjectiv}, we summarized the mean subjective scores of our method and the state-of-the-art (SOTA) virtual staining methods across these three dimensions in a heatmap. Scores of 0--2 were classified as unsatisfactory, while scores of 3--5 were classified as satisfactory. We calculated Fleiss' Kappa to assess the inter-rater consistency among the pathologists. In the MIST dataset, the Fleiss' Kappa for structural fidelity is 0.656, for visual quality is 0.628, and for diagnostic usability is 0.818. In the IHC4BC dataset, the Fleiss' Kappa values are 0.849 for structural fidelity, 0.847 for visual quality, and 0.930 for diagnostic usability, indicating a high level of consistency among the evaluators' subjective assessments.}

\section{Disscussion}

\subsection{ SSIM/PSNR is a good metric for Virtual Staining?}

\begin{figure}
    \centering
    \includegraphics[width=0.8\linewidth]{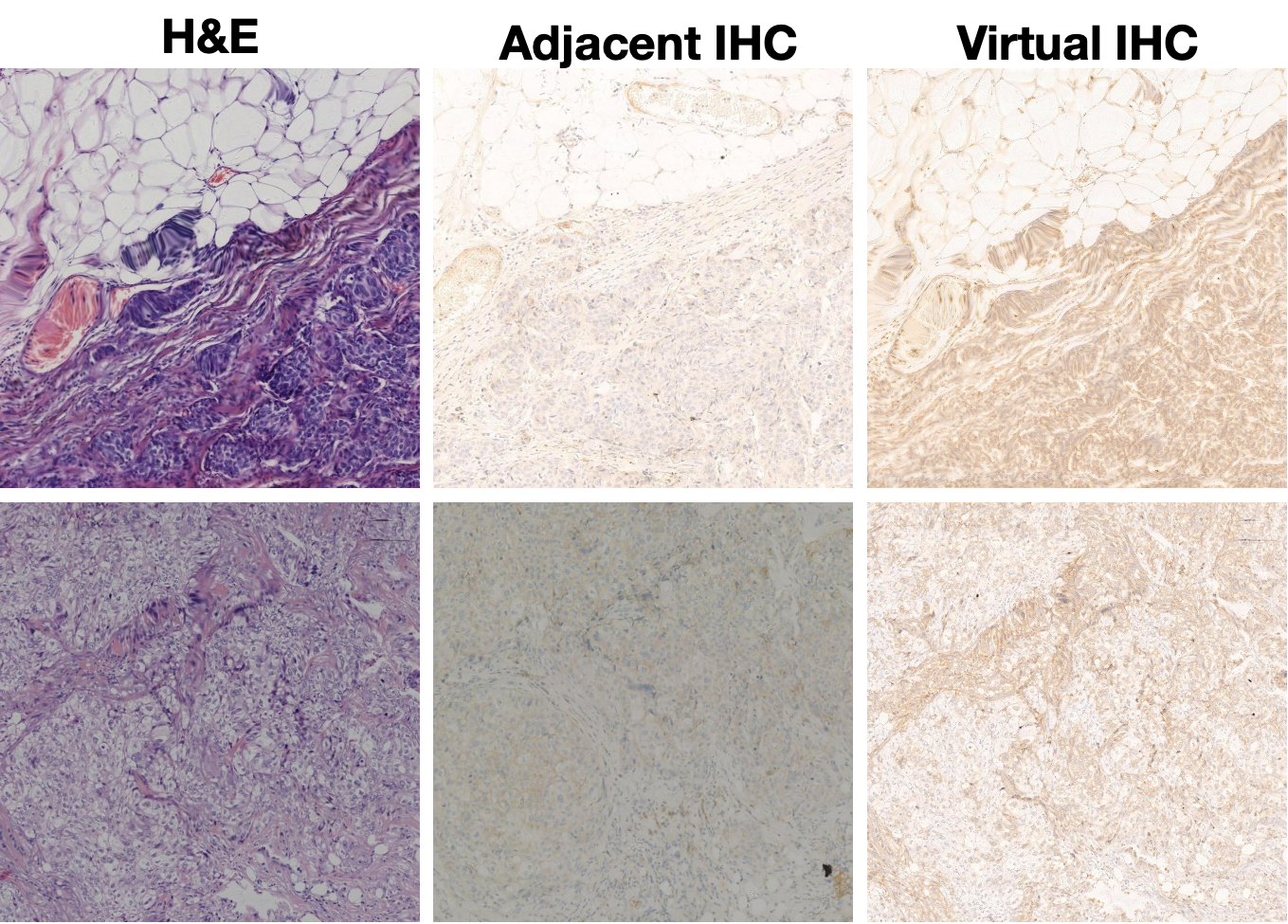}
    \caption{{The heatmap below represents the subjective evaluation scores of pathologists for Immunohistochemistry (IHC) images. These scores were assigned based on their assessment of diagnostic clinical value.}}
    \label{fig:disscusion1}
\end{figure}

% \begin{figure}
%     \centering
%     \includegraphics[width=\linewidth]{ood.pdf}
%     \caption{Real-world clinical data were collected for cross-center performance validation, demonstrating our method effectiveness on out-of-distribution (OOD) data. First row is virtual IHC and Second row is real IHC.}
%     \label{fig:ood}
% \end{figure}

{SSIM and PSNR are traditional image quality evaluation metrics that primarily measure pixel-level similarity and reconstruction quality. However, they face several limitations when applied to pathological images. The variability in brightness and contrast caused by imaging conditions, such as aperture size and light intensity, can lead to significant fluctuations, making strict brightness consistency across images difficult to achieve. As SSIM and PSNR rely on such consistency, their evaluation results can often be misleading in pathological scenarios. Furthermore, these metrics neglect critical pathological features like nuclear morphology and staining intensity, which are essential for accurate diagnosis, and therefore fail to capture high-level pathological semantics. They also struggle with the spatial heterogeneity inherent in weakly paired datasets, being overly sensitive to positional mismatches, which are often irrelevant to diagnostic outcomes. Finally, SSIM and PSNR focus solely on pixel-level similarity, disregarding the tissue structure and pathological semantics required to evaluate the contextual and diagnostic relevance of pathological images. As shown in Figure~\ref{fig:disscusion1}, the widely-used public dataset BCI\cite{ref_18} exhibits noticeable brightness variations.  Consequently, when brightness fluctuates, the SSIM metric, as well as pixel-level metrics like PSNR, may exhibit significant changes. However, such changes do not necessarily indicate errors in generating the staining intensity of positive regions; instead, they merely reflect variations in brightness. }

{Currently, most studies focus on using SSIM/PSNR to evaluate virtual staining results, while largely overlooking the unavoidable brightness variations commonly encountered in clinical settings. This is fundamentally different from the assessment of natural images. 
As \cite{ma2025generativeaimisalignmentresistantvirtual} illustrates, slight image rotation results in pixel-level distance changes. We attempted to adjust the brightness of the generated results on the BCI dataset based on the adjacent slices. Before global brightness adjustment, the evaluation metrics were as follows: SSIM = 0.3874 and PSNR = 18.8571. After applying global brightness adjustment, the metrics improved significantly to SSIM = 0.39717 and PSNR = 22.1215. However, despite the noticeable improvement in these metrics, the diagnostic significance of the generated images remained unchanged. Pixel-level metrics struggle to evaluate virtual staining results due to the imperfect registration of weakly paired data. Inspired by \cite{histDist}, we propose Path-FID, which replaces the InceptionV3 backbone of traditional FID with CONCH \cite{lu2024visual}, a model pretrained on billion-scale pathological data. Path-FID captures multi-level pathological semantics and tissue structures, making it robust to varying pairing rates. As shown in Table \ref{p-fid1} and Table \ref{p-fid2}, Path-FID demonstrates strong consistency in evaluating content and pathological semantics, validating its suitability for virtual staining tasks.}

\begin{table*}[tb]
    \centering
    \caption{The quantitative comparison of pathological relevance metrics and Path-FID on the MIST dataset.}
    \label{p-fid1}
    \renewcommand\arraystretch{1.2}
    \setlength{\tabcolsep}{1.5mm}{
    \resizebox{\linewidth}{!}{
    	{\begin{tabular}{l|ccccc|ccccc|ccccc|ccccc}
    		\toprule[1.0pt]
    		\multirow{2}{*}{Method}  
            & \multicolumn{5}{c|}{Ki67} 
            & \multicolumn{5}{c|}{ER} 
            & \multicolumn{5}{c|}{HER2} 
            & \multicolumn{5}{c}{PR} \\ 
            \cmidrule(lr){2-21}
            & $R_p$ & $R_c$ & $R_{avg}$ & IoD & P-FID 
            & $R_p$ & $R_c$ & $R_{avg}$ & IoD & P-FID 
            & $R_p$ & $R_c$ & $R_{avg}$ & IoD & P-FID 
            & $R_p$ & $R_c$ & $R_{avg}$ & IoD & P-FID \\ 
            
            \midrule[0.5pt]
            
            PyramidP2P  
            & 60.69 & 10.25 & 35.47 & -2.2201 & 291.7394 
            & 71.17 & 16.29 & 43.73 & -5.5017 & 188.5554 
            & 66.87 & 10.32 & 38.59 & -5.8272 & 165.1759 
            & 81.72 & 11.04 & 0.578 & -5.5563 & 198.0107\\

            ASP  
            & 56.63 & 85.91 & 71.27 & -1.8027 & 41.2966 
            & 63.53 & \textbf{86.40}  & 74.96 & -3.1462 & 36.8592 
            & 64.40 & \textbf{88.22} & \underline{76.31} & -4.7957 & 51.87  
            & 61.03 & \textbf{87.65} & 74.34 & -5.1900 & 42.1095 \\

            PPT  
            & 55.28 & 59.66 & 51.97 & -2.5072 & 101.3484 
            & 61.47 & 54.79 & 58.13 & -4.5220 & 135.7866 
            & 50.02 & 61.93 & 55.98 & -4.6784 & 88.6038  
            & 65.30 & 61.10 & 63.20 & -6.2238 & 97.3034 \\

            TDKStain  
            & \textbf{76.97} & 8.20 & 42.58 & \textbf{-0.1888} & 68.1517 
            & \textbf{86.54} & 12.78 & 49.66 & -2.6782 & 68.1517
            & 79.98 & 8.84 & 44.41 & -1.8509 & 84.7010
            & 85.43 & 11.32 & 48.38 & -3.2448 & 96.0378 \\

            UMDST  
            & 37.61 & 0.25 & 18.93 & -0.787 & 96.7640 
            & 46.19 & 82.80 & 64.00 & -2.9506 & 111.9784 
            & 15.98 & 82.93 & 49.46 & \textbf{0.1598} & 92.8557  
            & 52.16 & 83.30 & 67.73 & -2.5224 & 177.1798 \\

            SIMGAN  
            & 64.68 & \textbf{81.67} & 73.17 & -0.6221 & 19.3284
            & 72.28 & 84.26 & \underline{78.26} & -1.4809 & 22.6989 
            & 61.48 & 85.98 & 73.37 & \underline{-1.4421} & \underline{23.7088}  
            & \underline{85.16} & 85.76 & \underline{85.55} & \textbf{-0.2687} & \underline{26.9999} \\

            \cellcolor{gray!20}Ours   
            & \cellcolor{gray!20} \underline{73.87} & \cellcolor{gray!20} \underline{81.63} & \cellcolor{gray!20} \textbf{77.75} & \cellcolor{gray!20} \underline{-0.268} & \cellcolor{gray!20} \textbf{17.5884}
            & \cellcolor{gray!20} \underline{84.07} & \cellcolor{gray!20} \underline{86.06} & \cellcolor{gray!20} \textbf{85.07} & \cellcolor{gray!20} \textbf{-0.5109} & \cellcolor{gray!20} \textbf{21.0212}  
            & \cellcolor{gray!20} \textbf{80.08} & \cellcolor{gray!20} \underline{86.63} & \cellcolor{gray!20} \textbf{83.35} & \cellcolor{gray!20} -1.6788 & \cellcolor{gray!20} \textbf{22.5829} 
            & \cellcolor{gray!20} \textbf{87.37} & \cellcolor{gray!20} \underline{87.03} & \cellcolor{gray!20} \textbf{87.20} & \cellcolor{gray!20} \underline{-2.070} & \cellcolor{gray!20} \textbf{25.0436} \\

            \midrule[0.5pt]
    	\end{tabular}}}}
\end{table*}

\begin{table*}[tb]
    \centering
    \caption{The quantitative comparison of pathological relevance metrics and Path-FID on the IHC4BC dataset.}
    \label{p-fid2}
    \renewcommand\arraystretch{1.2}
    \setlength{\tabcolsep}{1.5mm}{
    \resizebox{\linewidth}{!}{
    	{\begin{tabular}{l|ccccc|ccccc|ccccc|ccccc}
    		\toprule[1.0pt]
    		\multirow{2}{*}{Method}  
            & \multicolumn{5}{c|}{Ki67} 
            & \multicolumn{5}{c|}{ER} 
            & \multicolumn{5}{c|}{HER2} 
            & \multicolumn{5}{c}{PR} \\ 
            \cmidrule(lr){2-21}
            & $R_p$ & $R_c$ & $R_{avg}$ & IoD & P-FID 
            & $R_p$ & $R_c$ & $R_{avg}$ & IoD & P-FID 
            & $R_p$ & $R_c$ & $R_{avg}$ & IoD & P-FID 
            & $R_p$ & $R_c$ & $R_{avg}$ & IoD & P-FID \\ 
            
            \midrule[0.5pt]
            
            PyramidP2P  
            & \textbf{89.84} & 7.12 & 48.48 & -0.7704 & 77.6012 
            & \underline{80.45} & 17.11 & 48.78 & -1.6351 & 73.9984
            & 72.79 & 17.68 & 45.24 & -8.0986 & 89.7197  
            & \textbf{66.28} & 15.74 & 41.01 &  -1.1295 & 71.9115 \\

            ASP  
            & 69.11 & \textbf{81.43} & 75.22 & 0.2612 & 57.6038 
            & 75.66 & \textbf{90.04} & \underline{75.65} & \underline{-0.2174} & 26.8832 
            & 27.24 & \textbf{94.76} & 61.00 & -9.9207 & 74.0689  
            & 15.59 & 87.64 & \underline{51.62} & -1.2598 & \underline{61.4121} \\

            PPT  
            & 81.60 & 63.38 & 72.49 & 0.9100 & 58.8053
            & -6.11 & 46.85 & 20.37 & 0.8693 & 102.7235 
            & 65.39 & 72.17 & 68.78 & -8.8844 & 58.1593
            & 6.66 & 67.01 & 36.83 & -1.1287 & 99.7219 \\

            TDKStain  
            & 81.17 & 4.54 & 45.86 & -0.6335 & 86.0319 
            & 68.50 & 19.16 & 43.83 & -0.8196 & 66.3562
            & \underline{75.58} & 15.28 & 45.43 & -6.645 & 74.1557 
            & 16.46 & 15.19 & 15.82 & -1.1883 & 108.7895 \\

            UMDST  
            & 67.45 & 18.97 & 43.21 & 1.0701 & 143.4910 
            & 26.62 & 82.58 & 54.60 & -1.5868 & 164.6007 
            & 48.78 & 82.01 & 65.40 & -4.1051 & 149.1963  
            & 15.53 & 82.31 & 48.92 & \textbf{-0.0899} & 176.2453 \\

            SIMGAN  
            & 83.19 & 80.41 & \underline{81.80} & 0.1395 & 52.3105 
            & 76.31 & 85.27 & 80.79 & 2.0767 & 31.2988 
            & 65.38 & 88.42 & \underline{76.90} & -0.1793 & 83.5756  
            & 9.10 & 81.37 & 45.23 & \underline{-0.3719} & \textbf{50.4651} \\

            \cellcolor{gray!20}Ours   
            & \cellcolor{gray!20} \underline{84.14} & \cellcolor{gray!20} \underline{80.33} & \cellcolor{gray!20} \textbf{82.23} & \cellcolor{gray!20} \textbf{0.009} & \cellcolor{gray!20} \textbf{16.4687}
            & \cellcolor{gray!20} \textbf{82.25} & \cellcolor{gray!20} \underline{86.03} & \cellcolor{gray!20} \textbf{78.47} & \cellcolor{gray!20} \textbf{0.0870} & \cellcolor{gray!20} \textbf{21.4066} 
            & \cellcolor{gray!20} \textbf{75.76} & \cellcolor{gray!20} \underline{89.21} & \cellcolor{gray!20} \textbf{81.98} & \cellcolor{gray!20} -7.7342 & \cellcolor{gray!20} \textbf{35.8315}  
            & \cellcolor{gray!20} \underline{36.57} & \cellcolor{gray!20} \underline{87.03} & \cellcolor{gray!20} \textbf{61.80} & \cellcolor{gray!20} 3.2447 & \cellcolor{gray!20} 130.6920 \\

            \midrule[0.5pt]
    	\end{tabular}}}}
\end{table*}

\subsection{Downstream Analysis}
{Following~\cite{varghese2014ihc}, we categorized the IHC images into different levels based on optical density thresholds, which were initially set and later adjusted by pathologists. Specifically, the HER2 dataset was divided into four levels: 0–500, 500–2000, 2000–5000, and 5000+. For ER and PR staining, the data was divided into two levels: 0–1000 and 1000+, while Ki67 was divided into 0–2000 and 2000+. These levels were used to train a ViT classification model as a downstream task for validation.}

{Although TDKStain achieved the best performance in grading results, as shown in Table \ref{cls_1} and Table \ref{cls_2}, its poor content consistency significantly limits its potential for practical applications. In contrast, the grading performance of our method across multiple datasets highlights its advantage in maintaining pathological semantic consistency, which is critical for clinical applicability.}
%%%%%%%%%%%%%%%%%%%%%%%%%%%%
\begin{table}[htb]%\normalsize %
    \centering
    \caption{{The accuracy of virtual staining state-of-the-art (SOTA) methods for breast cancer grading on the MIST dataset.}}
    \label{cls_1}
    % \vspace{-.15cm}
    %\renewcommand{\arraystretch}{1}
    \renewcommand\arraystretch{1.2} %4.2
    \setlength{\tabcolsep}{1.5mm}{
    \resizebox{\linewidth}{!}{
    	{\begin{tabular}{l|cc|cc|cc|cc}
    		\toprule[1.0pt]
    	\multirow{2}{*}{Method}  &\multicolumn{2}{c|}{Ki67} &\multicolumn{2}{c|}{ER} &\multicolumn{2}{c|}{HER2}&\multicolumn{2}{c}{PR}\\ \cmidrule(lr){2-9}&Recall($\uparrow$)   &F1-Score ($\uparrow$)  &Recall($\uparrow$)   &F1-Score  ($\uparrow$) &Recall($\uparrow$)   &F1-Score 
        &Recall   &F1-Score \\ 
            % \midrule[0.5pt]
            % \multicolumn{16}{c}{mvtec-3dad} \\
            \midrule[0.5pt]
            
            PyramidP2P  &0.842 &0.7735 &0.695& 0.6633 &0.587 &0.5820  &0.705&0.7044\\
            ASP &0.759 &0.7767 &0.623& 0.62405 &0.378 &0.3961  &0.637&0.6339\\
            PPT  &0.8 &0.8057 &0.5460 & 0.5607 &0.45 &0.4320  &0.642&0.5155\\
            TDKStain &0.6400 &0.6812 &\textbf{0.8050} & \textbf{0.8087} &\textbf{0.7285} &\textbf{0.6520}  &\textbf{0.7790}&\textbf{0.7799}\\
            % MDCL  &\underline{00.0} &-- / -- &00.0 & 1.1 / 1.1 &00.0 &3.3 / 3.3  &00.0&11.1 / 3.3\\
            
            % \midrule[0.5pt]
            UMDST &0.4091 &0.4091 & 0.5840 & 0.5962 &0.4200 &0.5336  &0.5770 &0.5733\\
            SIMGAN &0.8280 &0.8247 &0.6780 &0.6775 &0.6168 &0.5000 &0.7050&0.7019\\
            % GramGAN &00.0 &-- / -- &00.0 & 1.1 / 1.1 &00.0 &3.3 / 3.3  &00.0&11.1 / 3.3\\
            % \midrule[0.5pt]
            \cellcolor{gray!20}Ours   &\cellcolor{gray!20} \textbf{0.8510  }&\cellcolor{gray!20}\textbf{0.8404}
             &\cellcolor{gray!20}\underline{0.7240} &\cellcolor{gray!20}\underline{0.7290}  
            &\cellcolor{gray!20}\underline{0.6827} &\cellcolor{gray!20}\underline{0.6080}  &\cellcolor{gray!20}\underline{0.7310}
            &\cellcolor{gray!20}\underline{0.7319}\\

            \midrule[0.5pt]
    \end{tabular}}}}
    
\end{table}

\begin{table}[htb]%\normalsize
    \centering
    \caption{{The accuracy of virtual staining state-of-the-art (SOTA) methods for breast cancer grading on the IHC4BC dataset.}}
    \label{cls_2}
    % \vspace{-.15cm}
    %\renewcommand{\arraystretch}{1}
    \renewcommand\arraystretch{1.2} %4.2
    \setlength{\tabcolsep}{1.5mm}{
    \resizebox{\linewidth}{!}{
    	{\begin{tabular}{l|cc|cc|cc|cc}
    		\toprule[1.0pt]
    	\multirow{2}{*}{Method}  &\multicolumn{2}{c|}{Ki67} &\multicolumn{2}{c|}{ER} &\multicolumn{2}{c|}{HER2}&\multicolumn{2}{c}{PR}\\ \cmidrule(lr){2-9}&Recall($\uparrow$)   &F1-Score ($\uparrow$)  &Recall($\uparrow$)   &F1-Score  ($\uparrow$) &Recall($\uparrow$)   &F1-Score 
        &Recall   &F1-Score \\ 
            % \midrule[0.5pt]
            % \multicolumn{16}{c}{mvtec-3dad} \\
            \midrule[0.5pt]
            
            PyramidP2P  &0.686 &0.6527 &0.497& 0.4679 &0.505 &0.4545  &0.64&\underline{0.6068}\\
            ASP &\underline{0.799} &\underline{0.7975} &0.547 & 0.4958 &0.256 &0.2369  &0.559&0.4989\\
            PPT  &0.71 &0.6815 &0.505 & 0.4892 &0.6080 &0.378  &\underline{0.642}&0.5155\\
            TDKStain &0.774 &0.7635 &\underline{0.759} & \underline{0.7507} &0.471 &\underline{0.4887}  &\textbf{0.759}&\textbf{0.7507}\\
            % MDCL  &\underline{00.0} &-- / -- &00.0 & 1.1 / 1.1 &00.0 &3.3 / 3.3  &00.0&11.1 / 3.3\\
            
            % \midrule[0.5pt]
            UMDST &0.7890 &0.7825 &0.573 & 0.5701 &0.412 &0.4218  &0.599&0.6055\\
            SIMGAN &0.7650 &0.7506 &0.6960 & 0.6513 &\underline{0.6448} &\textbf{0.5110}  &0.6010&0.5982\\
            % GramGAN &00.0 &-- / -- &00.0 & 1.1 / 1.1 &00.0 &3.3 / 3.3  &00.0&11.1 / 3.3\\
            % \midrule[0.5pt]
            \cellcolor{gray!20}Ours   &\cellcolor{gray!20} \textbf{0.8110  }&\cellcolor{gray!20}\textbf{0.8109}
             &\cellcolor{gray!20}\textbf{0.8100} &\cellcolor{gray!20}\textbf{0.8076}  
            &\cellcolor{gray!20}\textbf{0.6504} &\cellcolor{gray!20}0.4490  &\cellcolor{gray!20}0.3877
            &\cellcolor{gray!20}0.4450\\

            \midrule[0.5pt]
    \end{tabular}}}}
    
\end{table}

%[Network G] Total number of parameters : 7.838 M
\begin{table}[!ht]
    \centering
    \caption{{Evaluate the efficiency of model on the MIST ER dataset.}}
    % \resizebox{\columnwidth}{!}{
    \begin{tabular}{cccc}
        \toprule
         Method   & $R_p$ & Parameters(M) & Inference time  \\
        \midrule
         CycleGAN &  46.94 & 11.383 & 0.0253s \\
         PyramidP2P & 71.17  & 11.383   & 0.0103s  \\ % 
         UMDST   & 45.18 & 10.204 & 0.0892s \\ % 
         % GramGAN   & - & - & - \\ % 
         ASP   & 63.53 & 7.838 & 0.0114s \\ % 
         TDKStain   & 86.54 & 45.593 & 0.0198s \\ % 
         SIMGAN   & 72.28 & 7.838  & 0.0119s  \\ % 
         Ours    & 84.07 & 7.838   & 0.0115s  \\ % 
        \bottomrule
    \end{tabular}%}
    \label{runtimetb}
\end{table}

\subsection{Efficient deployment improvements}
{Due to the limited computational resources in hospitals and the billion-pixel nature of whole-slide imaging (WSI) data, the clinical deployment of virtual staining faces certain challenges and adjustments. As shown in Table~\ref{runtimetb}, our method achieves comparable parameter efficiency and processing speed to lightweight models. On whole-slide images, our method processes a slide at 40x magnification in approximately 2.51 minutes and at 20x magnification in about 37.7 seconds. However, these results are based on online cloud servers, where computational power is significantly higher than that of deployment environments. Current research has shown that Spiking Neural Networks (SNNs)\cite{ref_snn_1,ref_snn_2,ref_snn_3,ref_snn_4,ref_snn_5,ref_snn_6,ref_snn_7}, which operate through discrete spike events in a biologically inspired manner, can significantly reduce the computational resource demands of neural networks. Exploring ways to convert existing models into SNNs holds significant clinical implications.}

\section{Conclusion}
This paper proposes USIGAN, a novel IHC virtual staining method that completely eliminates weakly paired terms. By employing cyclic optimal transport and intra-batch optical density consistency, USIGAN ensures both content and pathological semantic consistency in virtual IHC staining. However, it remains a one-to-one mapping virtual staining model. In future work, we plan to extend the characteristics of optimal transport to multi-domain virtual staining, enabling the model to learn differential features between various staining types within a shared framework.

% \section{Declaration of competing interest}
% The authors declare that they have no known competing financial interests or personal relationships that could have appeared to influence  the work reported in this paper.

\section{Ackowledgements}
This work was supported in part by the National Natural Science Foundation of China (No. 62271475), Ministry of Science and Technology’s key research and development program (2023YFF0723400), Shenzhen-Hong Kong Joint Lab on Intelligence Computational Analysis for Tumor lmaging (E3G111), and the Youth Innovation Promotion Association CAS (2022365).

% % ========= 返稿加 =====
% \section{Data avaliability}
% MIST Dataset can be requested on: \url{ https://link.springer.com/chapter/10.1007/978-3-031-43987-2_61} and IHC4BC dataset can be requested on: \url{https://ihc4bc.github.io/}

% \bibliography{references}
{\small % 在这里调整字体大小
\printbibliography
}

\begin{IEEEbiography}[{\includegraphics[width=1in,height=1.25in,clip,keepaspectratio]{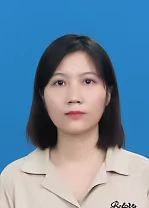}}]{Yue Peng}
 is currently pursuing a PhD at the University of Chinese Academy of Sciences and the Shenzhen Institute of Advanced Technology, Chinese Academy of Sciences. Her research interests include computational pathology, medical image segmentation, and medical image generation.\end{IEEEbiography}
\begin{IEEEbiography}[{\includegraphics[width=1in,height=1.25in,clip,keepaspectratio]{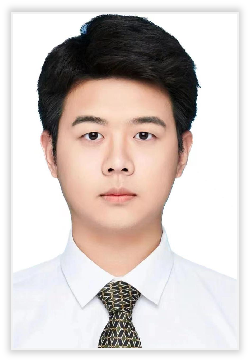}}]{Bing Xiong}
recived the B.S. degree in the school of Wuhan University of Technology,Hubei,China,in 2023. He is currently pursuing the M.S degree with ShenZhen Institues of Advanced Technology,University of the Chinese Academy of Sciences. His research intersts include computational pathology and image generation.\end{IEEEbiography}
\begin{IEEEbiography}[{\includegraphics[width=1in,height=1.25in,clip,keepaspectratio]{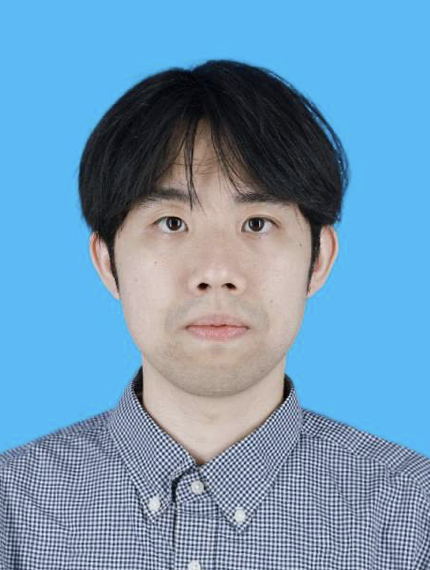}}]{Fuqiang Chen}
recived the B.S. degree in the school of  Microelectronics and Communication Engineering, Chongqing University, Chongqing, China, in 2022, the M.S. degree in the Shenzhen Institute of Advanced Technology, Chinese Academy of Sciences and University of Chinese Academy of Sciences, in 2025. His research interests include computational pathology.\end{IEEEbiography}
\begin{IEEEbiography}[{\includegraphics[width=1in,height=1.25in,clip,keepaspectratio]{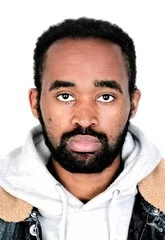}}]{Deboch Eybo Abera}
 received the B.S. degree in Electronics Information Engineering from the University of Electronic Science and Technology of China in 2021, and the M.S. degree in Information and Communication Engineering from the same university in 2023. He is currently pursuing a Ph.D. degree in Pattern Recognition and Intelligent Systems at the Shenzhen Institute of Advanced Technology, Shenzhen, China. His research interests include digital pathology, multi-modality microscopy image processing etc.\end{IEEEbiography}
\begin{IEEEbiography}[{\includegraphics[width=1in,height=1.25in,clip,keepaspectratio]{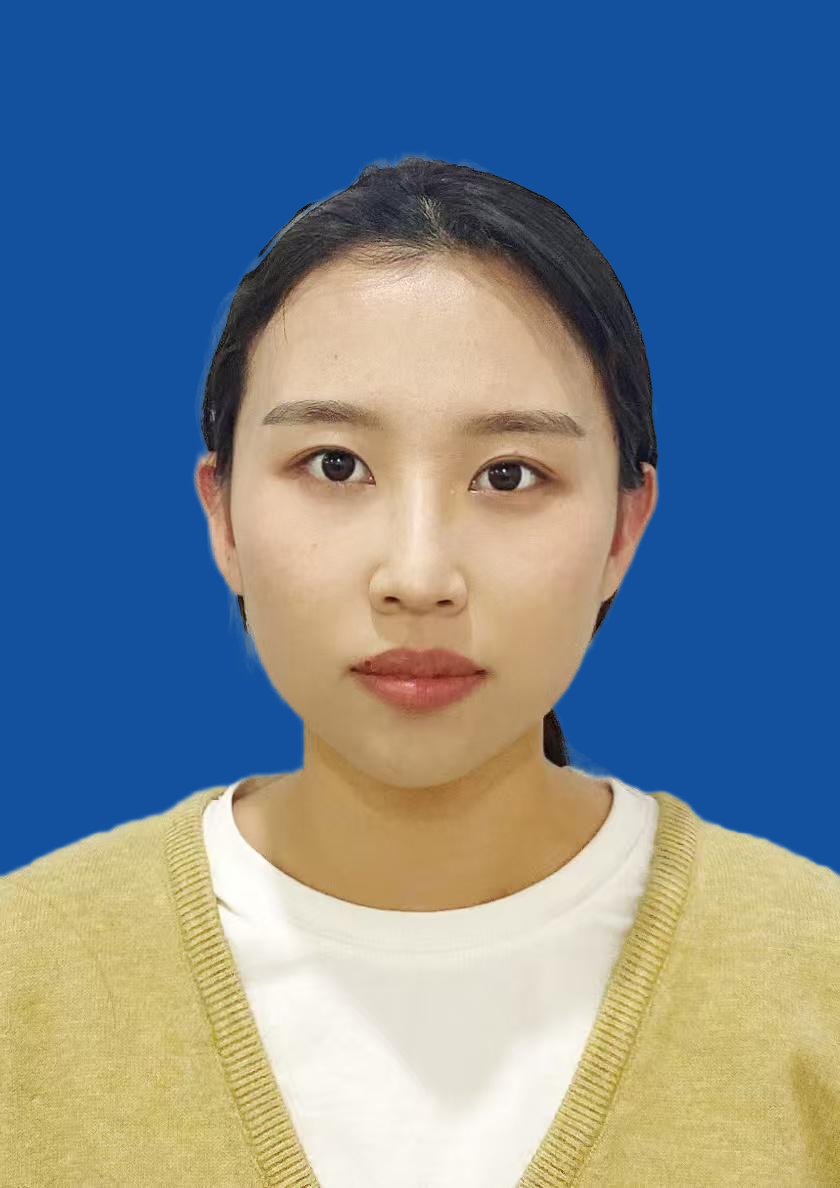}}]{RanRan Zhang}
recived the B.S. degree in the
school of Physics and Physical Engineering,Qufu Normal University,Qufu, China, in 2017, the Ph.D. degree in the school of Information Science and Engineering, Shandong University, Qingdao,
China,in 2023.She is currently working as a postdoctoral fellow at the Shenzhen Institute of Advanced Technology, Chinese Academy of Sciences. Her research interests include computational pathology and deep learning.\end{IEEEbiography}
\begin{IEEEbiography}[{\includegraphics[width=1in,height=1.25in,clip,keepaspectratio]{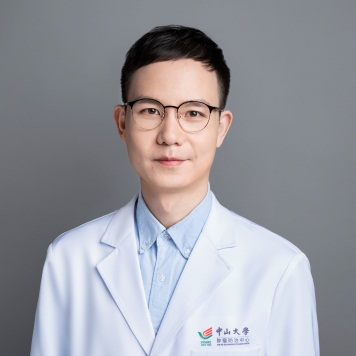}}]{Wanming Hu}
received the B.M. degree in Clinical Medicine from CSMU, Changsha, China (2010) and the M.M. and M.D. degrees in Pathology from Xiangya School of Medicine, CSU, Changsha, China (2013) and Southern Medical University, Guangzhou, China (2020). He is currently an associate Chief Physician, currently serves as Young Editorial Board Member of Chinese Journal of General Surgery, with primary research focusing on central nervous system tumors and head/neck tumors (pathological diagnosis, molecular subtyping, immune microenvironment, AI pathology, etc.).\end{IEEEbiography}
\begin{IEEEbiography}[{\includegraphics[width=1in,height=1.25in,clip,keepaspectratio]{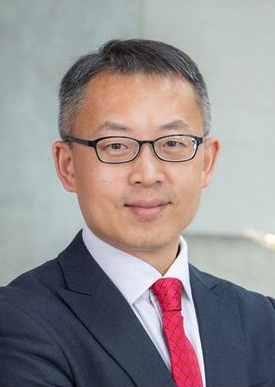}}]{Jing Cai}
recived B.S degree in Lanzhou University,Lanzhou,China, and M.S. degree in University of Georgia,USA. and the Ph.D. degree from University of Virginia,USA. he is currently the head professor of   Department of Health Technology \& Informatics, HK PolyU.His research interst in  Novel medical imaging and image preprocessing techniques.\end{IEEEbiography}
\begin{IEEEbiography}[{\includegraphics[width=1in,height=1.25in,clip,keepaspectratio]{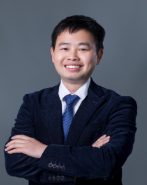}}]{Wenjian Qin}
is currently the associate professor of the Institute of Biomedical and Health Engineering at Shenzhen Institute of Advanced Technology, Chinese Academy of Sciences. Dr. Qin obtained his Ph.D. in Pattern Recognition and Intelligent System from the University of Chinese Academy of Sciences.  Dr. Qin has been engaged in the research of  multi-modality medical imaging with machine learning and computer vision, which explores the new computing theories and methods of the learning-based algorithm in clinical diagnosis and treatment.\end{IEEEbiography}

\end{document}